\documentclass[12pt]{article}

\usepackage[T1]{fontenc}
\usepackage[utf8]{inputenc}
\usepackage{hyperref}      
\usepackage{url}            
\usepackage{booktabs}       
\usepackage{amsfonts}       
\usepackage{nicefrac}       
\usepackage{microtype}      
\usepackage{graphicx}
\usepackage{amsmath,amssymb,amsthm}
\usepackage{float}
\usepackage{caption}
\usepackage{mathtools}
\usepackage{graphicx}
\usepackage{setspace}
\usepackage{lmodern}
\usepackage{xcolor}
\usepackage{enumitem}
\usepackage{placeins}

\usepackage{geometry}
\numberwithin{equation}{section}
\theoremstyle{plain}
\newtheorem{theorem}{Theorem}[section]
\newtheorem{lemma}[theorem]{Lemma}

\theoremstyle{definition}
\newtheorem{definition}{Definition}

\newtheorem{assumption}{Assumption}
\newtheorem{prop}{Proposition}
\usepackage{times}

\usepackage{graphicx} 
\usepackage[numbers]{natbib}

\usepackage{algorithm}
\usepackage{algorithmic}
\usepackage{bbm}
\usepackage[caption=false]{subfig}

\def\cont{{(c)}}

\def\argmin{\mathop{\rm arg\,min}}
\newcommand{\reals}{{{\mathbb{R}}}}

\newcommand{\expect}{{{\mathbb{E}}}}

\newcommand{\ind}[1]{{{\mathbb{I}_{\{#1\}}}}}

\newcommand{\X}{{\cal X}}

\def\ie{{\em i.e.,~}}
\def\cf{{\em cf.,~}}
\def\eg{{\em e.g.,~}}

\title{Network Estimation from Point Process Data}

\author{Benjamin Mark$^{1,4}$, Garvesh Raskutti$^{2,4}$, and Rebecca Willett$^{3,4}$\\
$^1$Department of Mathematics, $^2$Department of Statistics, \\ $^3$Department of Electrical and Computer Engineering \\
$^4$Wisconsin Institute for Discovery\\
University of Wisconsin-Madison 
}

\begin{document}
\maketitle

\begin{abstract}Consider observing a collection of discrete events within a network that reflect how network nodes influence one another. Such data
are common in spike trains recorded from biological neural
networks, interactions within a social network, and a variety of other settings. Data of this form may be modeled as self-exciting point processes,
in which the likelihood of future events depends on the past events.
This paper addresses the
problem of estimating self-excitation parameters and inferring the underlying functional network structure from
self-exciting point process data.  Past work in this area was limited by strong assumptions which are addressed by the novel approach here. Specifically, in this paper we (1) incorporate \emph{saturation} in a point process model which both ensures stability and models non-linear thresholding effects; (2) impose general low-dimensional structural assumptions that include sparsity, group sparsity and low-rankness that allows bounds to be developed in the high-dimensional setting; and (3) incorporate long-range memory effects through moving average and higher-order auto-regressive components. Using our general framework, we provide a number of novel theoretical guarantees for high-dimensional self-exciting point processes that reflect the role played by the underlying network structure and long-term memory. We also provide simulations and real data examples to support our methodology and main results.\end{abstract}

\section{Introduction}
In a variety of settings, our only glimpse of a network’s structure is through the lens of discrete time series
observations. For instance, in a social network, we may observe a time series of members’ activities,
such as posts on social media. In electrical systems, cascading chains of power failures reveal
critical information about the underlying power distribution network. During epidemics, networks
of computers or of a population are reflected by the time at which each node becomes infected. In
biological neural networks, firing neurons can trigger or inhibit the firing of their neighbors, so that
information about the network structure is embedded within spike train observations.

This paper focuses on estimating the \emph{influence network} which models the extent to which one node’s activity stimulates or inhibits
activity in another node. For instance, the network structure may indicate who is influencing whom
within a social network \cite{raginsky_OCP,silva:pami,BertozziHawkes,dynamicMirrorDescent,HellerHawkes,zhouZhaSongHawkes}, the connectivity of neurons \cite{brown2004multiple,onlineHawkesNeuro,colemanConvexPoint,SmithBrownStateSpace,HinneHeskes2012,
    DingSchroeder2011,spikesPillow,lindermanneuro}, interactions among financial instruments \cite{chavez2012high,ait2010modeling,linderman2014discovering}, how power
failures may propagate across the power grid \cite{ertekin2015reactive},
or patterns of criminal activity and military engagements
\cite{BertozziHawkes,HellerHawkes,egesdal2010statistical,lewis2012self,linderman2014discovering}. The interactions between nodes are thus
critical to a fundamental understanding of the underlying functional network structure and accurate
predictions of likely future events. 

Learning the influence network presents a number of challenges both in
terms of formulating the model and developing suitable theory and
methodology. First, in the applications described above the number of
network nodes is typically large relative to the length of time they
are observed, making the network parameter
\emph{high-dimensional}. Furthermore, the most natural model in these
settings are multivariate {\em self-exciting point processes (SEPPs)}. While
empirical work has demonstrated the efficacy of SEPP models in various applications (\cf
\cite{egesdal2010statistical,chavez2012high,lewis2012self,linderman2014discovering,ertekin2015reactive}),
little is known about the statistical properties of these
estimators. In this paper, we formulate a model and provide a general
framework for estimating network parameters in discrete-time
high-dimensional SEPP models.

Let $M$ denote the number of nodes in the network and $T$ the number of time intervals over which we collect data. We observe $X_{t,m}$, the number of events at node $m$ during time period $t$, for $m= 1,\ldots,M$ and $t = 1,\ldots, T$. We model these counts as
$$X_{t,m} \sim \mbox{Poisson}(\lambda_{t,m})$$ where the logarithm of $\lambda_{t,m}$ is a function of the previous counts of events in the network and the interactions between nodes. For a simple example, we might have $\log \lambda_{t,m} = \sum_{m'=1}^m A_{m,m'} X_{t-1,m'}$. However, a fundamental challenge associated with SEPP models is that they can be highly unstable: due to the exponential link function, the counts can diverge even when the interactions $\{A_{m,m'}\}$ are small.  In \cite{stability} the authors give extensive justification for the interest in these models from a neuroscience perspective, but also show how learned model parameters can result in generative models that are highly inconsistent with physiological measurements.  Existing statistical learning bounds for SEPP models \cite{hall2016inference} guarantee stability by assuming all network interactions are inhibitory. 

A major contribution of this work is learning guarantees for SEPPs without restrictive assumptions on the structure of the network or types of interactions among nodes.
We will address stability issues by introducing saturation effects on the rate parameter $\lambda_{t,m}$. Saturated SEPP models were recently described in application-driven work without theoretical guarantees \cite{ertekin2015reactive}.  In contrast, this work aims to derive statistical learning guarantees for saturated point processes.  

We study a fairly general class of saturated SEPPs whose parameters can be estimated via regularized maximum likelihood estimation.  We assume that the number of possible interactions between nodes (\ie graph edges) $M^2$ is large relative to the number of time points $T$, but that the network has an underlying low-dimensional structure that can be promoted via regularization. The question we address in our theory is how many time points $T$ are needed to guarantee a desired level of statistical accuracy in terms of the number of nodes $M$, the underlying network structure, the regularizer used, and the type of saturation effects introduced?

\subsection{Relationship to Prior Work}

A number of works have studied {\em linear} SEPPs (where $\lambda_{t,m}$ is a linear function of past events, in contrast to {\em log-linear} models, where $\log \lambda_{t,m}$ is a linear function of past events) from a theoretical perspective.  Examples include works on the Hawkes process \cite{clipped_hawkes, bacry, patriciaHawkes, hansen2012lasso, bacry2}.  In a multivariate Hawkes process setting, one frequently aims to learn the excitation matrix characterizing interactions within the network.  In \cite{etesami2016learning} the authors establish that learning the excitation matrix is sufficient for learning the directed information graph of the process.  The linear Hawkes process is frequently studied under an assumption there are no inhibitory interactions, although recent work \cite{clipped_hawkes} was able to incorporate both inhibitory and stimulatory interactions.  Prior work on learning parameters in discrete high-dimensional time series
models requires linearity or Gaussianity assumptions (\cf \cite{BasuMichailidis15}) which do not hold in our model.  

In contrast, we study log-linear SEPPs.  Prior works have demonstrated the empirical value of log-linear SEPPs \cite{ppglm, ppglm3} and these models are frequently used in the neuroscience community \cite{stability}.  Moreover, log-linear point process models can be advantageous from the perspective of optimization \cite{hybridPoisson}  and naturally allow for inhibitory interactions.  However, log-linear SEPPs can not easily model stimulatory interactions while maintaining stability, and incorporating stimulatory interactions is a major contribution of this paper.

There is limited work on learning rates for log-linear SEPPs, and much of it is only applicable in the setting where $M$ is small relative to
$T$ \cite{fokianos2011log}.  The most related work is our recent work \cite{hall2016inference} which considers a special case of our SEPP along with a sparsity assumption on the network and applies in the high-dimensional setting. This prior work is limited since the model only considers
recent memory, sparsity regularization, and assumes only
inhibitory influences to ensure stability and
learnability.

\subsection{Main Contributions}

Our paper makes the following major contributions.

\begin{itemize}

\item We provide a general upper bound (Theorem~\ref{Thm:bound}) for developing theoretical guarantees for estimating SEPPs and build on the analysis in~\cite{hall2016inference} in three significant ways. First, we incorporate saturation effects in our model by using a thresholding function in order to ensure stability, and account for these effects in our theory. Second, we provide learning rates for a class of processes which incorporate longer-range dependence effects in a variety of ways, improving upon~\cite{hall2016inference} which only considers first-order auto-regressive models.  Finally, we allow for several different regularization choices corresponding to various prior beliefs about the structure of the network. 
  
\item We apply our general upper bound to a number of different processes and regularization schemes. For processes with longer-range dependence, we prove that a restricted eigenvalue condition holds for the ARMA$(1,1)$ and AR$(2)$ models in Lemmas~\ref{Lem:ar1} and~\ref{Lem:two_basis} respectively. 
\item In terms of regularization schemes, we consider strict sparsity, group sparsity and low-rank regularization and provide three novel guarantees stated in Theorems \ref{cor:sparse}, \ref{cor:gp} and \ref{cor:rnk}. All our mean-squared error bounds match the optimal bounds in the independent case up to log factors.
\item A thorough simulation study in Section VI provides support for our theoretical mean-squared error bounds and also examines parameters associated with the magnitude of the entries of $A^*$ and clipping thresholds.  
\item We further demonstrate the practical benefits of our regularized likelihood framework on three real data examples. The first involves modeling the interplay between crime events in different neighborhoods of Chicago, the second modeling connections between different neurons in the brain within a rat during sleep and wake states, and the third involving meme-tracker data in social networks. The three examples illustrate the advantages of using different regularizers. 
\item Finally, we show that our SEPP framework can be viewed as a discretization of the widely studied Hawkes process, and discuss some advantages of considering point processes in discrete time.
\end{itemize}

\subsection{Notation}
For a matrix $A$, we let $a_{m.}$ denote the $m^{\rm th}$ row of $A$ and $a_{.m}$ denote the $m^{\rm th}$ column of $A$. We then let $\|a_{m.}\|_{1+}$ denote the sum of the positive entries of $a_{m.}$ and $\|a_{m.}\|_{1-}$ denote the absolute value of the sum of the negative entries of $a_{m.}$, so that 
$$\|a_{m.}\|_1 = \|a_{m.}\|_{1+} + \|a_{m.}\|_{1-}.$$ 
Given a norm $\|\cdot\|_{\cal R}$ on a real vector space, we let $\|\cdot\|_{\cal R^\ast}$ denote its dual norm defined by $$\|v\|_{\cal R^\ast}=\sup_{\|u\|_{\cal R} \leq 1} <u,v>$$ where $<\cdot, \cdot>$ denotes the standard inner product.
Throughout the paper, we work with mixed norms $$\|A\|_{p,q}=(\sum_m \|a_{m.}\|_q^p)^{\frac{1}{p}},$$ as well as the nuclear norm $$\|A\|_\ast=\sum_{i=1}^M \sigma_i(A)$$ where $\sigma_i(A)$ denotes the $i$th singular value of $A$, and the operator norm $$\|A\|_{op}=\sup_{\|x\|_2\leq 1} \|Ax\|_2.$$  The Frobenius norm, denoted by $\|\cdot \|_F$, is a special case of the $\|\cdot\|_{p,q}$ norm with $p=q=2$.

Finally, we let $\|A\|_0$ denote the number of nonzero elements of a matrix $A$.

\section{Model Formulation}

In this section, we present a class of SEPPs and discuss how saturation effects can be included in order to ensure stability.  Recall that $X_{t,m}$ denotes the number of events from node $m$ during time period $t$.  To start, consider the following model:
\begin{equation}X_{t+1,m} \sim \mbox{Poisson}(\lambda_{t+1,m})\end{equation}
\begin{equation}\log(\lambda_{t+1,m})=\nu_m+\sum_{s=1}^{t} \sum_{m'=1}^M h_{m,m'}[t-s] X_{s,m'}.\label{eq:initial_model}\end{equation}
Here the logarithm of the rate for $X_{t+1,m}$ is linear in all the previous observations.  For each node $m'$ in the network, that node's count $X_{s,m'}$ at time $s$ is scaled by an influence function $h_{m,m'}$ evaluated at $t-s$.  The influence function $h_{m,m'}$ describes the relationship between nodes $m$ and $m'$.  As in \cite{clipped_hawkes}, we assume each influence function can be written as the linear combination of $K$ known basis functions $\{\phi_k\}_{k=1}^K$, \ie 
$$h_{m,m'}[t]=\sum_{k=1}^K a_{m,m',k} \phi_k[t].$$
Hence estimating the network structure amounts to estimating the  matrix $A^\ast \in \reals^{M \times MK}$ where
the $m^{\rm th}$ row of $A^\ast$ is $\begin{pmatrix}\begin{pmatrix}
a_{m,m',k} 
\end{pmatrix}_{m'=1}^M \end{pmatrix}_{k=1}^K$
It will be convenient to rewrite  ~\eqref{eq:initial_model} in matrix-vector form as 
\begin{equation} \log(\lambda_{t+1})=\nu+A^\ast g(\mathcal{X}_t), \label{eq:model} \end{equation} 

where $\mathcal{X}_t=[X_1,\ldots , X_t]$ denotes the history of the process up to time $t$ and $g(\mathcal{X}_T) \in \reals^{MK \times 1}$ is the vector defined as follows. For $k=1,\ldots,K$, let \begin{equation}g_k(\mathcal{X}_T) :=
\label{eq:gk}
\begin{bmatrix}
\sum_{s=1}^{T} X_{s,1} \phi_k [T-s]\\ \sum_{s=1}^{T} X_{s,2} \phi_k [T-s]\\ \vdots \\ \sum_{s=1}^{T} X_{s,M} \phi_k [T-s]
\end{bmatrix}
\end{equation}
and
\begin{equation}g(\mathcal{X}_T) := \label{eq:gxt}
\begin{bmatrix} 
g_1(\mathcal{X}_T) \\ g_2(\mathcal{X}_T) \\ \vdots \\ g_K(\mathcal{X}_T)
\end{bmatrix}.
\end{equation}

A number of commonly studied discrete time models can be realized in this manner.  We briefly mention two which are discussed further in Section IV.  As a first example, $K=1$ and $\phi[t]=\alpha^t$ corresponds to an autoregressive moving average ARMA$(1,1)$ process.  When $K=p$ and $\phi_k[t]=\ind{k=t}$ where $\ind{B} := {\small \begin{cases}1, &B \mbox{ true}\\ 0, & \mbox{otherwise} \end{cases}}$ is the indicator function, we recover the AR(p) process. This second example shows the value in assuming that $h_{m,m'}$ is in the span of a collection of basis functions, rather than just one. Allowing for multiple basis functions allows us to study processes which incorporate higher order effects in more sophisticated ways than would be possible with only one basis function.

We let $\nu_{\min}$ and $\nu_{\max}$ be upper and lower bounds on the constant offset parameter $\nu_m$ in \eqref{eq:initial_model} 
and we assume that $A^\ast$ lies within a set $\mathcal{A}$ which we define as follows. Let $a_{\max}$ be an upper bound on $\|a_{m.}^\ast\|_{1+}$ and similarly let $a_{\min}$ be an upper bound on $\|a_{m.}^\ast\|_{1-}$. We let $\mathcal{A}$ denote the set of $M \times MK$ matrices with $\|a_{m.}\|_{1-} \leq a_{\min}$ and $\|a_{m.}\|_{1+} \leq a_{\max}$ for all $m$. 
With the assumption that $A^\ast \in \mathcal{A}$ we can search for an estimate $\widehat{A}$ of $A^\ast$ over the bounded set $\mathcal{A}$.

\subsection{Saturation} As discussed in the introduction, point process models along the lines of \eqref{eq:model} are widely used to describe count data in a variety of applications.  However, due to instability issues inherent to SEPPs of this form, these models can be highly unstable and lead to unbounded counts. Hence, pure SEPPs make poor generative models (c.f., \cite{stability}) and are difficult to understand theoretically without making overly restrictive assumptions about $A^\ast$ (c.f, \cite{hall2016inference}).  We will address this problem by introducing saturation effects to the vector $g(\mathcal{X}_t)$ defined in Equation \eqref{eq:gxt}.  The application focused work \cite{ertekin2015reactive} introduced saturated SEPPs, but to the best of our knowledge, this is the first work to study the theoretical properties of saturated models.  To address stability issues we adjust the definition of $g_k(\mathcal{X}_T)$ in  \eqref{eq:gk} to the following: 
\begin{equation}g_k(\mathcal{X}_T)=
\begin{bmatrix}
\sum_{s=1}^{T} \min(X_{s,1},{\tilde U}) \phi_k [T-s]\\[6pt] \sum_{s=1}^{T} \min(X_{s,2},{\tilde U}) \phi_k [T-s]\\ \vdots \\ \sum_{s=1}^{T} \min(X_{s,M},{\tilde U}) \phi_k [T-s]
\label{eq:gk2}
\end{bmatrix}.
\end{equation}
That is, each past count which exceeds some threshold ${\tilde U} \geq 1$ gets clipped to ${\tilde U}$.  Further, we assume that 
$$\sum_{s=1}^\infty \phi_k[s]  \leq \tau < \infty$$
for each basis function, so that each entry of $g(\mathcal{X}_t)$ in  \eqref{eq:gxt} is bounded by $${\tilde U} \sum_{s=1}^T \phi_k[s] \leq \tau {\tilde U} =: U.$$  
In other words, with clipping we have $\|g(\mathcal{X}_t)\|_\infty \leq U$, guaranteeing the stability of our process. 

In particular, this allows us to define the maximum and minimum Poisson rate from which each observation can be drawn.  We denote the maximum and minimum rates by
\begin{subequations}
\label{eq:rmax}
\begin{align}
R_{\max}=&\exp\left(\nu_{\max}+a_{\max}U\right)\\
R_{\min}=&\exp\left(\nu_{\min}-a_{\min}U\right).
\end{align}
\end{subequations}

Throughout this paper, we take $\min(.,{\tilde U})$ to be our saturation function for simplicity.  However, our theory extends to other saturation functions provided that the function is bounded, which is crucial for our analysis. The details are provided in Proposition \ref{prop:saturation} in the appendix.  

While this framework has advantages, a central question we need to address is how departing from the standard SEPP framework and incorporating non-linear saturation effects change our estimation errors.

\subsection{Regularized optimization formulation}

In the high-dimensional setting, the number of potential pairwise interactions, $M^2$, is large relative to the number of time periods, $T$, making standard maximum likelihood optimization techniques unsuitable.  Instead, we assume some prior knowledge on the parameter $A^\ast$, which can be incorporated in estimation via a regularization term $\|.\|_{\mathcal{R}}$.
Specifically, we consider the estimator
\begin{align}
\hat A =& \argmin_{A \in \mathcal{A}} \sum_{t=0}^\top \sum_{m=1}^M
\exp\left(\nu_{m}+a_{m.}^\top g(\mathcal{X}_t)\right) -X_{t+1,m}(\nu_{m}+a_{m.}^\top g(\mathcal{X}_t))+ \lambda \|A\|_\mathcal{R} \label{eq:estimator} 
\end{align}
where the first two terms of $\eqref{eq:estimator}$ are the negative log-likelihood of the observed data given $A$. We discuss various choices of regularization $\|.\|_{\mathcal{R}}$ in the next section. Note that the optimization problem in \eqref{eq:estimator} is convex. Further, it can
easily be generalized to unknown $\nu$; we omit this discussion here for simplicity of presentation.

\section{Statistical learning bounds}
\label{sec:main}

\subsection{Decomposable Regularizers}
Our learning bounds apply to general decomposable regularizers introduced in \cite{Neg10}. Given a subspace 
$\overline{\mathcal{M}} \subseteq \mathbb{R}^{M \times MK}$, we define its orthogonal complement as
$$\overline{\mathcal{M}}^\perp=\{v \in \mathbb{R}^{M \times MK} | <u,v>=0 \text{ for all } u \in \overline{\mathcal{M}}\}. $$
Given a normed vector space $(\mathbb{R}^{M \times MK}, \|\cdot 
\|_\mathcal{R})$  and subspaces $\mathcal{M} \subseteq \overline{\mathcal{M}} \subseteq \mathbb{R}^{M \times MK}$, we say $\mathcal{R}$ is a decomposable regularizer with respect to $(\mathcal{M},\overline{\mathcal{M}}^\perp)$ if for $A \in \mathcal{M}$ and $B \in \overline{\mathcal{M}}^\perp$ we have $$\|A+B\|_\mathcal{R}=\|A\|_\mathcal{R}+\|B\|_\mathcal{R}.$$  This definition encompasses widely-studied regularizers including the $l_1$ norm, nuclear norm, and the group sparsity inducing $\|\cdot\|_{1,2}$ norm. We refer the reader to \cite{Neg10} for more details and intuition.  While working in this general framework allows us to incorporate a wide variety of prior beliefs about the structure of our network, a fundamental question we need to address is how the specific choice of regularizer effects our learning rates.  Due to the temporal dependence and non-linearities in SEPP models, deriving learning rates for various decomposable regularizers requires us to leverage martingale concentration inequalities.

\subsection{Assumptions}In Section II we presented a class of SEPPs which depends on a choice of basis functions, and a general RMLE procedure which depends on a choice of regularization penalty.  In this section, we introduce four assumptions which are needed for our theoretical guarantees. We then give examples where we show that for certain choices of basis functions $\{\phi_1,\ldots \phi_K\}$ and regularizers $\|.\|_{\mathcal{R}}$ of interest the assumptions hold with high probability.  

Our first assumption depends on the basis functions but is not related to the choice of regularizer.

\begin{assumption}[Restricted Eigenvalue]
\label{A2}
There exists some $\omega>0$ and $p \in \mathbb{N}$ such that smallest eigenvalue of $\mathbb{E}[g(\mathcal{X}_t)g(\mathcal{X}_t)^\top |\mathcal{X}_{t-p}]$ is lower bounded by $\omega$ for all $t$.
\end{assumption}

Assumption \ref{A2} is analogous to various restricted eigenvalue conditions in other works.   However, in much of the literature, one needs to lower bound the eigenvalues of a sensing matrix whose columns are assumed to be independent.  Dependence introduced in our autoregressive model makes this a more complex condition to verify.  In past work on sparse autoregressive inference (\eg \cite{BasuMichailidis15}), restricted eigenvalue conditions have been framed in terms of a stationary covariance matrix.

Informally, the value of $\omega$ measures the strength of the intertemporal dependence of our process.  If our network and basis functions are structured such that strong long-range dependencies exist, then the smallest eigenvalue can be near zero, leading to a poor bound on the error $\|\widehat{A}-A^\ast\|_F^2$.  

The RE condition must also account for the level of clipping in our process: if the network is so stimulatory that most observations are clipped, then the matrix $\mathbb{E}[g(\mathcal{X}_t)g(\mathcal{X}_t)^\top|\mathcal{X}_{t-p}]$ 
will be nearly singular and $\omega$ will be close to zero.  Thus, to come up with an acceptable bound on $\omega$, we need to establish that our network is well-behaved enough that most observations will be unclipped.  In other words, our theory suggests that introducing non-linear saturation effects will not ruin our ability to infer the structure of our network, provided that our network is not too stimulatory and is usually stable without clipping. A further discussion of the intuition behind the RE condition is provided in Example 1.

\medskip

Next, we present assumptions which need to be verified in terms of the regularizer used.  Recall that we assume the regularizer $\|\cdot \|_{\cal R}$ used in Equation \eqref{eq:estimator} is decomposable with respect to the pair of subspaces $(\mathcal{M},\overline{\mathcal{M}}^\perp)$. 

\begin{assumption}[Subspace Compatibility]\label{A3} There exists a constant $\Psi(\overline{\mathcal{M}})$ satisfying 
 $$\sup_{A \in \overline{\mathcal{M}}} \frac{ \|A\|_\mathcal{R}}{\|A\|_F} \leq \Psi(\overline{\mathcal{M}}).$$
 
\end{assumption}

Assumption \ref{A3} is a subspace compatibility condition as in \cite{Neg10}, which controls how large the Frobenius norm can be relative to the $\mathcal{R}$ norm on the subspace $\overline{\mathcal{M}}$.  

\begin{assumption}[Cone Row Sparsity]\label{A4}
 Let $A_{\overline{\mathcal{M}}}$ and $ A_{\overline{\mathcal{M}}^\perp}$ denote the projections of a matrix $A$ onto the subspaces $\overline{\mathcal{M}}$ and $\overline{\mathcal{M}}^\perp$ respectively.  Define 
\begin{align*}
\mathcal{B}'_{\mathcal{R}}=\big\{&A \in \mathbb{R}^{M \times MK} :  \|A_{\overline{\mathcal{M}}^\perp}\|_\mathcal{R} \leq 3 \|A_{\overline{\mathcal{M}}}\|_\mathcal{R} \text{ and } \|A\|_F=1\big\}.
\end{align*}
Then there exists a constant $\mu_{\cal R}$ such that $$\sup_{B \in \mathcal{B}'_{\mathcal{R}}} \|B\|_{2,1}^2 \leq \mu_\mathcal{R}.$$ 
\end{assumption}

Assumption \ref{A4} corresponds to assuming some notion of row sparsity on the error matrix $\widehat{A}-A^\ast=: \triangle$.  It is needed to apply the empirical process techniques from \cite{hall2016inference}.

\begin{assumption}[Deviation Bound]\label{A5}
 Let $$\epsilon_{t,m}=X_{t+1,m}-\exp(\nu_m+a_{m.}^\top g(\mathcal{X}_t)),$$  then there exists a constant $\lambda < \infty$ such that $$\left \Vert \frac{1}{T} \sum_{t=1}^T \epsilon_t g(\mathcal{X}_t)^\top \right \Vert_{\mathcal{R}^\ast} \leq \frac{\lambda}{2}.$$

\end{assumption}

Assumption \ref{A5} is similar to deviation bound conditions found in the literature.  Due to the temporal dependence across observations, we must use martingale concentration inequalities under various norms in order to verify it.

\subsection{General result}

Provided our process and estimation procedure satisfy Assumptions \ref{A2}-\ref{A5} for reasonable constants, we can guarantee the learnability of our model.  

\begin{theorem} \label{Thm:bound}
Assume $(X_t)_{t=1}^T$ is generated by \eqref{eq:model} and satisfies Assumptions \ref{A2}-\ref{A5} and assume $A^\ast$ is estimated according to \eqref{eq:estimator} with a regularizer $\|\cdot \|_\mathcal{R}$ that is decomposable with respect to the subspaces $(\cal M, \overline{\cal M}^\perp)$.  Then 
$$\|\widehat{A}-A^\ast\|_F^2 \leq \frac{36 p \Psi(\overline{\mathcal{M}}) \lambda^2}{ R_{\min}^2 \omega^2}$$ with probability at least $1-\frac{2}{M^2}$ for $$T \geq  \frac{128p^2U^4\mu_{\mathcal{R}}^2 \log M}{\omega^2}$$ for constants $C,c>0$ which are independent of $M,T$ and $\Psi(\overline{\mathcal{M}})$.  
\end{theorem}

Theorem ~\ref{Thm:bound} is a direct consequence of Theorem 1 in \cite{Neg10} combined with Theorem 1 in \cite{hall2016inference}.  Specifically, \cite{Neg10} gives Theorem \ref{Thm:bound} in a general decomposable regularizer setup under a restricted strong convexity (RSC) assumption, which in our language states that the error $\triangle=: \widehat{A}-A^\ast$ satisfies $$\frac{1}{T} \sum_t \sum_m (\triangle_{m.}^\top g(\mathcal{X}_t))^2 \geq k \|\triangle\|_F^2$$ for some $k>0$.  Due to the fact that our process is neither linear nor Gaussian, many techniques, \eg \cite{BasuMichailidis15} \cite{rnk}, used to establish an RSC condition directly are unworkable in our setting.  Instead, we use similar techniques to Theorem 1 in \cite{hall2016inference} which uses empirical process results to turn the RSC assumption into the restricted eigenvalue (RE) condition in Assumption 1.

\section{Examples}

In order to use Theorem \ref{Thm:bound}, we need to prove that the four assumptions hold for basis functions and regularizers of interest.  First, we show that Assumption \ref{A2} is satisfied for different point process models.  Second we show Assumptions \ref{A3}-\ref{A5} are satisfied for a class of regularizers.  Finally, we combine the results from this section with Theorem \ref{Thm:bound} to give overall learning rates for ARMA$(1,1)$ and AR$(2)$ processes under different regularization schemes.

Recall that the constants $R_{\max}$ and $R_{\min}$ from  \eqref{eq:rmax} denote the maximum and minimum Poisson rate for each observation.

\subsection{Specific Point Process Models}

\paragraph{Example 1: ARMA$(1,1)$ process}

\sloppypar First-order autoregressive moving average (ARMA$(1,1)$) point process models have been studied in a variety of settings \cite{lewis2012self, kasspaper}.  Moreover, the corresponding continuous time model is one of the most frequently studied point process models \cite{ait2010modeling,PointProcesses}. Consider the following saturated ARMA$(1,1)$ model with memory parameter $\alpha \in [0,1)$:
\begin{align}X_{t+1}\sim &\mbox{ Poisson}(\lambda_{t+1})\nonumber \\
 \log(\lambda_{t+1})=&\nu+A^\ast \min(X_t,{\tilde U})+\alpha\log(\lambda_{t}).\label{eq:arma} 
 \end{align}
Algebraic manipulation shows that \eqref{eq:arma} is a special case of  \eqref{eq:model} with $K=1$ basis function corresponding to $\phi[t]=\alpha^t$.  Here $\alpha$ is a memory parameter which captures the strength of the long-range dependence in the process, and $\|g(\mathcal{X}_t)\|_\infty \leq \frac{\tilde U}{1-\alpha}=U$ 
so that 
\begin{align*}
R_{\max}=&\exp\left(\nu_{\max}+\frac{a_{\max}\tilde{U}}{1-\alpha}\right)\\ R_{\min}=&\exp\left(\nu_{\min}-\frac{a_{\min}\tilde{U}}{1-\alpha}\right).
\end{align*}

An AR(1) process, corresponding to  \eqref{eq:arma} where $\alpha=0$, was considered in~\cite{hall2016inference}. However, due to the inherent instability of SEPPs without saturation, the authors were forced to assume $a_{\max} = 0$.

\begin{lemma}
\label{Lem:ar1}

Suppose $(X_t)_{t=1}^T$ is generated according to ~\eqref{eq:arma}.  Then Assumption \ref{A2} is satisfied with $$\omega=\min\left(\frac{1}{2}R_{\min},\kappa\right)$$ where $\kappa$ is a constant depending on $\tilde{U}$, $\alpha$ and $a_{\max}$ but independent of $M$. 
\end{lemma}

The proof of Lemma \ref{Lem:ar1} requires us to account for the effects of clipping.  We show
that finding a lower bound on the eigenvalues of $\expect[g(\X_t)g(\X_t)^\top|\X_{t-p}]$ can be reduced to finding a
lower bound on $\mbox{Var}(\min(X_{t,m},\tilde{U})|\mathcal{X}_{t-1})$, which simplifies the calculation since we rely on first-order dependence. Since we need to
construct a lower bound on $\mbox{Var}(\min(X_{t,m},\tilde{U})|\mathcal{X}_{t-1})$ we
consider the two cases when the variance will be smallest. 

In particular, if $X_{t,m} \sim \mbox{Poisson} (R_{\min})$, then its variance will be small because the variance and mean of a Poisson random variable are equal. 
Specifically, when $X_{t,m} \sim \mbox{Poisson} (R_{\min})$ we lower bound the variance of $\min(X_{t,m},\tilde{U})$ by $\frac{1}{2}R_{\min}$. 

On the other hand, when $X_{t,m} \sim \mbox{Poisson} (R_{\max})$ and  $R_{\max}$ is large relative to $\tilde{U}$, then $\min(X_{t,m},\tilde{U})$ is likely to be
$\tilde{U}$ (clipped), so again the variance will be small. We lower bound the variance by the constant $\kappa$ that is the variance of a Bernoulli random variable, where one outcome corresponds to a Poisson random variable $Z \sim \mbox{Poisson}(R_{\max})$  exceeding $\tilde U$ (clipped) and the other outcome corresponds to $Z < \tilde U$. 

One of these two worst case scenarios will give an absolute lower bound on the variance. In both cases we construct a lower bound on the variance independent of $M$. 
Note that $\omega$ increases with $R_{\min}=\exp(\nu_{\min}-\frac{a_{\min}\tilde{U}}{1-\alpha})$ so $\omega$ grows inversely with $\alpha$.  In other words, as the long range memory of the process increases, Lemma \ref{Lem:ar1} suggests that network estimation becomes more difficult.  This is consistent with prior work \cite{BasuMichailidis15}.

The value $\kappa^{-2}$ may be viewed as a proxy for the rate of clipping, and the appearance of $\kappa$ in Lemma \ref{Lem:ar1} illustrates a tradeoff associated with clipped models.  On one hand, clipping ensures a stable process.  However, as clipping increases, it also increases the temporal dependencies among observations, leading to a smaller $\kappa$ and larger error bound.   

\begin{figure}[ht]
\label{fig:heat_map}
	\centering
	\includegraphics[width=.9\linewidth]{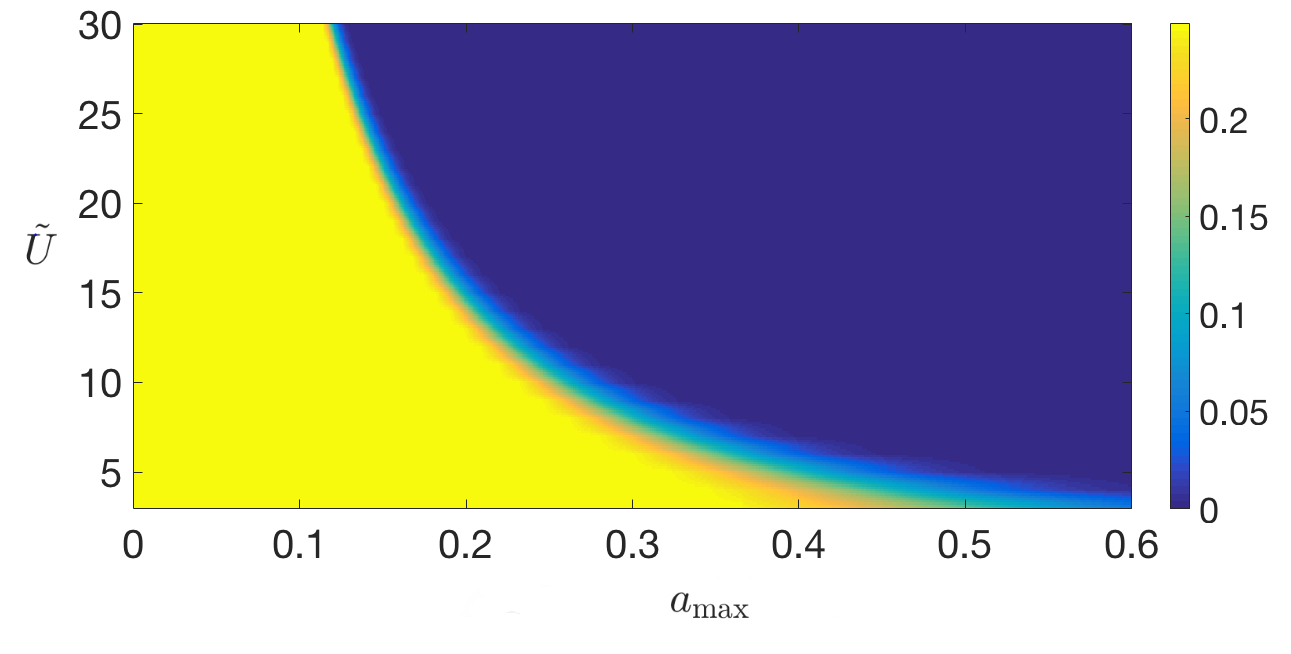}
	\caption{Colors indicate the value of $\kappa$ for a given $(a_{\max},\tilde{U})$ pair.  Note that due to our exponential link functions, elements of $A^*$ above one would be unreasonably excitatory for many networks, and our process can be significantly stimulatory even with coefficients well below one.}
	\label{fig:heatmap}
\end{figure}

 In Figure \ref{fig:heatmap}, we fix $\alpha=0$ and
get a sense of the value of $\kappa$ for varying $a_{\max}$ and $\tilde{U}$. (Recall that larger $\kappa$ corresponds to a better-posed estimation problem.)  We see that for small $\tilde{U}$, $\kappa$ is not prohibitively small for a wide range of values of $a_{\max}$.
However, as $\tilde{U}$
increases the range of reasonable $a_{\max}$ decreases, and as $\tilde{U}$ approaches $\infty$, we approach the $a_{\max}=0$ setting from \cite{hall2016inference}.  To understand this trend, we consider a special case of \eqref{eq:arma} with $M=1$, $\nu=0$, $\alpha=0$, $A^\ast=\frac{3}{10}$ and $\tilde{U}=1000$. In this case, the process follows: $$X_{t+1} \sim \mbox{Poisson}\left(\exp\left(\frac{3 \min(X_t,1000)}{10}\right)\right).$$ Given a small $X_0$, this process will not be clipped for the first few observations, but eventually the process will diverge and reach the clipping threshold of $1000$. This will happen within the first $100$ observations with probability $\approx 1$.  Once an observation reaches $X_t \geq 1000$ and is clipped, the next observation will follow \begin{align*}X_{t+1} &\sim  \mbox{Poisson}\left (\exp \left (\frac{3 \min(X_t,1000)}{10}\right ) \right)\\&=\mbox{Poisson}(\exp(300)).\end{align*}  and so $X_{t+1}$ is virtually guaranteed to be clipped as well.  In other words, once we actually reach the clipping threshold, we enter a constant clipping regime which is reflected in the small value of $\kappa$ for $a_{\max}=.3$ and $\tilde{U}=1000$.

On the other hand, if $\tilde{U}=6$ our process follows $$X_{t+1} \sim \mbox{Poisson}\left (\exp \left(\frac{3\min(X_t,6)}{10}\right)\right).$$ Even when $X_t \geq 6$, the Poisson rate $\exp(\frac{18}{10})$ is approximately $6$, and so the next observation is reasonably likely to be unclipped and we do not enter the constant clipping loop as when $\tilde{U}=1000$.  In other words, over the long run we experience less clipping for smaller $\tilde{U}$, and thus $\kappa$ is larger for smaller $\tilde{U}$.

\paragraph{Example 2: AR$(2)$ process} As a second example, we consider an AR process with two time lags:
\begin{align} &X_{t+1}\sim \mbox{Poisson}(\lambda_{t+1})\\
 \log(&\lambda_{t+1})=\nu+A_1^\ast \min(X_t,\tilde{U})+A_2^\ast \min(X_{t-1},\tilde{U}).  \label{eq:ar2} \end{align} 
This process fits within the framework of  \eqref{eq:model} with two basis functions corresponding to: $\phi_1[t]=\ind{t=1}$ and $\phi_2[t]=\ind{t=2}$,   $\underbrace{A^\ast}_{M \times 2M}=[A_1^\ast, A_2^\ast]$. This example illustrates the benefit of considering a basis with more than one element to describe the influence functions $h_{m,m'}$.  A richer class of higher order models can be expressed with multiple basis functions.  
Under this setup, the maximum and minimum possible Poisson rates are 
$$R_{\max}=\exp(\nu_{\max}+a_{\max}\tilde{U})$$ and $$R_{\min}=\exp(\nu_{\min}-a_{\min}\tilde{U}).$$

Learning rates for high-dimensional linear AR(p) processes with Gaussian noise were studied in \cite{BasuMichailidis15}.  However, the techniques used in that work to prove a restricted eigenvalue condition relied heavily on the Gaussianity of the process.  We prove that the restricted eigenvalue condition in Assumption \ref{A2} holds for the AR$(2)$ process in Lemma \ref{Lem:two_basis}.

In order to state Lemma \ref{Lem:two_basis} we first need several definitions.  A node $m$ is said to be a \textit{parent} of node $m'$ if it influences $m'$ through $A_1^\ast$, while $m'$ is said to be a \textit{child} of $m$.  Furthermore, two nodes are said to be \textit{siblings} if they share a parent node.

\begin{lemma}
\label{Lem:two_basis}

Suppose $(X_t)_{t=1}^T$ is generated according to ~\eqref{eq:ar2}.  Let $\rho_{m}^{(p)}$ denote the number of parents of $m$, let $\rho_{m}^{(c)}$ denote the children of $m$ and let $\rho_{m}^{(s)}$ denote the number of siblings of $m$.  Then $$\lambda_{\min}(\mathbb{E}[g(\mathcal{X}_t)g(\mathcal{X}_t)^\top|\mathcal{X}_{t-2}]) \geq r_\rho>0$$ for a constant $r_\rho$ depending on $R_{\max}$, $R_{\min}$, $\rho_{m}^{(p)}$, $\rho_{m}^{(c)}$, $\rho_{m}^{(s)}$ but independent of $M$.

\end{lemma}

The constant $r_\rho$ scales inversely with $\rho_{m}^{(p)},\rho_{m}^{(c)}$, $\rho_{m}^{(s)}$ and $R_{\max}-R_{\min}$.  In the high dimensional setting, this means a sparsity assumption on $A^\ast$ is necessary for our bound to be useful.  

We prove Lemma \ref{Lem:two_basis} by showing that the matrix $\mbox{Cov}(g(\mathcal{X}_t)|\mathcal{X}_{t-2})$ is strictly diagonally dominant.  A matrix $B$ is said to be strictly diagonally dominant if there exists a constant $\omega>0$ such that $b_{i,i} -\sum_{j \not = i} |b_{i,j}|\geq \omega$ for all $i$, and the eigenvalues of a symmetric strictly diagonally dominant matrix are lower bounded by $\omega$.  With a sparsity assumption on $A^\ast$, almost all of the off diagonal terms in $\mbox{Cov}(\mathcal{X}_t)$ will be zero, and the remaining terms can be controlled with the techniques from Lemma \ref{Lem:ar1} and appropriate assumptions on the size of $R_{\min}$ and $R_{\max}$ relative to the sparsity constants $\rho_{m}^{(p)},\rho_{m}^{(c)},\rho_{m}^{(s)}$.  

\subsection{Regularization Examples}
\label{sec:regularization}
In this subsection, we verify Assumptions \ref{A3}-\ref{A5} under various regularization schemes.
\paragraph{Example 1: Element-wise Sparsity Regularization} We first explore sparsity regularization for these processes that accounts for
the sparsity of $A^\ast$ natural to many application domains.  For the remainder of the section, we assume $$\|A^\ast\|_0=s \ll M^2.$$ Sparse models of network structure encapsulate essential aspects of
many common statistical network models \cite{goldenberg2010survey}, and have connections to stochastic block
models, exponential random graph models, and various graphical models.
We consider the regularizer $$\|A\|_{1,1}=\sum_i \sum_j |a_{i,j}|$$ along with its dual $$\|A\|_{\infty,\infty}=\max_i \max_j |a_{i,j}|.$$
To see that $\|\cdot\|_{1,1}$ is decomposable we first define the set $$S= \{(i,j) \in \{1,\ldots, M\} \times \{1, \ldots, MK\} : A_{i,j}^{\ast} \not =0\},$$ and next define $$\mathcal{S}=\{s \in \mathcal{R}^{M \times MK} : s_{i,j}=0 \text{ for all } (i,j) \not \in S\}.$$ 
Then $\|\cdot \|_{1,1}$ is decomposable with respect to the pair $(\mathcal{S},\mathcal{S}^\perp)$.   

Note that the optimization problem corresponding to $\|\cdot \|_{1,1}$ regularization is convex
and can be solved with a variety of sparse regularization
solvers. Furthermore, it can trivially be parallelized across the rows
of $A$. 

\begin{lemma}
\label{Thm:Sparsity}
Suppose $(X_t)_{t=1}^T$ is generated according to  \eqref{eq:model} with $\|A^\ast\|_0=s$. Further, assume $A^\ast$ is estimated according to \eqref{eq:estimator} using $\|\cdot \|_{1,1}$ regularization.  Then 
\begin{enumerate}[label=(\alph*)]  
\item Assumption \ref{A3} is satisfied with $$\Psi(\overline{\mathcal{S}})=4\sqrt{s}.$$ 
\item Assumption \ref{A4} is satisfied with $$\mu_{(1,1)}=4\sqrt{s}.$$
\item Assumption \ref{A5} is satisfied with $$\|\frac{1}{T} \sum_{t=1}^T \epsilon_t g(\mathcal{X}_t)^\top\|_{\infty, \infty} \leq CR_{\max}\frac{\log^3(MT)}{\sqrt{T}}$$ with probability at least $1-\frac{1}{(MT)^c}$ for constants $C,c>0$ which are independent of $M,T$ and $s$.
\end{enumerate}
\end{lemma}

\paragraph{Proof Overview}
\begin{itemize}
\item To verify Assumption~\ref{A5}, we
rely on the fact that the Poisson rate can never exceed $R_{\max}$.
This allows us to bound the largest recorded observation by
$C\log(MT)$ with high probability. From here, we are in a position to
use martingale concentration inequalities developed in \cite{HoudreReynaudBouret03} to establish the deviation bound. 
\item Assumptions \ref{A3} and \ref{A4} are straightforward consequences of the relation between $l_1$ and $l_2$ norms.
\end{itemize}

Combining Lemma \ref{Thm:Sparsity}, Theorem \ref{Thm:bound} and the restricted eigenvalues results from the previous subsection gives overall bounds for sparse SEPPs which are applicable in the high-dimensional setting. 

\begin{theorem} (\textbf{Learning rates for $l_1$ regularization})
\label{cor:sparse}
Suppose $(X_t)_{t=1}^T$ follows the SEPP framework of 
  \eqref{eq:model} and $A^\ast$ is estimated using sparsity regularization.
 
\begin{enumerate}[label=(\alph*)]  
 \item If $(X_t)_{t=1}^T$ is generated according to the ARMA$(1,1)$ model in  \eqref{eq:arma} then
 $$\|\widehat{A}-A^\ast\|_F^2 \leq C \frac{R_{\max}^2}{R_{\min}^2 \min(\frac{1}{2}R_{\min},\kappa)^2} \frac{s \log^6(MT)}{T}$$ 
with probability at least $1-\frac{1}{(MT)^c}$ for $T,M$ satisfying $$T \geq 128U^4s\frac{\log M}{\min(\frac{1}{2}R_{\min},\kappa)^2}$$ for constants $C,c >0$ which are independent of $M,T$ and $s$.
\item If $(X_t)_{t=1}^T$ is generated according to the AR$(2)$ model in  \eqref{eq:ar2} then
 $$\|\widehat{A}-A^\ast\|_F^2 \leq C \frac{R_{\max}^2}{R_{\min}^2 r_\rho^2} \frac{s \log^6(MT)}{T}$$ 
with probability at least $1-\frac{1}{(MT)^c}$ for $T,M$ satisfying $$T \geq 128 U^4s\frac{\log M}{r_\rho^2}.$$ 

\end{enumerate}
\end{theorem}

The mean-squared error bound $\frac{s \log^6(MT)}{T}$ matches the minimax optimal rate in the independent case~\cite{RasWaiYu11} up to log factors. Theorem~\ref{cor:sparse} extends results in Hall et al.~\cite{hall2016inference} to ARMA(1,1) and AR(2) processes.

\paragraph{Example 2: Group Sparsity}

Group lasso regularization is a popular tool used to estimate a sparse parameter where one has prior knowledge on the structure of the sparsity (see \eg \cite{grouplasso} for more details). We consider a special case of group lasso regularization where the groups are the columns of the matrix.  Let $a_{.m}$ denote the $m^{\rm th}$ column vector of a matrix $A$.  Our structured sparsity assumption is that only $s_G \ll M$ columns of $A^\ast$ contain nonzero entries.

In terms of network structure, this means that only a small number of hub nodes have influence on other nodes in the network.  To estimate networks of this form, we consider $l_2$ penalization on the columns vectors, followed by $l_1$ penalization on the resulting $l_2$ norms.  In other words, we have $$\|A\|_G=\|A^\top\|_{1,2}=\sum_m \|a_{.m}\|_2.$$ 
The dual of this norm is $$\|A\|_{G^\ast}=\|A^\top\|_{\infty,2}=\max_m \|a_{.m}\|_2.$$
Let $$S_G=\{ i : a_{.i}^\ast \not =0\}.$$
Then $\|\cdot \|_G$ is decomposable with respect to the subspaces $$\mathcal{S}_G=\{ A : a_{.j}=0 \text{ for all } j \not \in S_G\}$$ and $$\mathcal{S}_G^\perp = \{ A: a_{.j}=0 \text{ for all } j  \in S_G\}.$$  We show Assumptions \ref{A3}-\ref{A5} hold in Lemma \ref{Thm:Group} below.

\begin{lemma}
\label{Thm:Group}

Suppose $(X_t)_{t=1}^T$ is generated according to  ~\eqref{eq:model} where only $s_G$ columns of $A^\ast$ contain nonzero entries.  Further, assume $A^\ast$ is estimated according to \eqref{eq:estimator} using $\|\cdot\|_G$ regularization.  Then 
\begin{enumerate}[label=(\alph*)]

\item Assumption \ref{A3} is satisfied with $$\Psi(\mathcal{S}_G)=4\sqrt{s_G}.$$
\item Assumption \ref{A4} is satisfied with $$\mu_{G}=16s_G.$$
\item Assumption \ref{A5} is satisfied with  $$\|\frac{1}{T} \sum_{t=1}^T \epsilon_t g(\mathcal{X}_t)^\top\|_{G^\ast} \leq CR_{\max} \log^2(MT)\sqrt{\frac{M}{T}}$$ with probability at least $1-\frac{1}{(MT)^c}$.
\end{enumerate}
\end{lemma}

\paragraph{Proof Overview}
\begin{itemize}
\item For Assumption \ref{A5} we construct a high probability bound on the $l_2$ norm of each column of our noise matrix and take a union bound over all the columns to get a final bound on the $\|\cdot \|_{\infty, 2}$ norm.  To bound the norm of each individual column, we rely on \cite{rakh} which provides martingale concentration inequalities for 2-smooth norms.  
\item For Assumption \ref{A4} we derive an error row sparsity constant which depends only on $s_G$ rather than $M$.  The $\|\cdot \|_{2,1}$ norm can be large relative to the Frobenius norm in cases where the matrix is row-dense.  In this case, the $l_1$ norm of each row can be on the order of $\sqrt{M}$ larger than the $l_2$ norm.  However, we only need to derive a compatibility constant on the cone $$\mathcal{B}_G=\{B \in \mathbb{R}^{M \times MK} : \|B_{\overline{\mathcal{S}_G}^\perp}\|_G \leq 3\|B_{\overline{\mathcal{S}_G}}\|_G\}.$$  Since elements of $\overline{\mathcal{S}_G}$ has at most $s_G$ nonzero entries in each row, we can think of all matrices in the cone $\mathcal{B}_{G}$ as being ``almost row sparse'' and so the $\|\cdot \|_{2,1}$ norm should not be $O(\sqrt{M})$ larger than the Frobenius norm on the cone.

\item Assumption \ref{A3} follows from the relationship between the $l_1$ and $l_2$ norms.
\end{itemize}
Combining Lemma \ref{Thm:Group}, Theorem \ref{Thm:bound} and the restricted eigenvalue conditions from the previous subsection gives the following result.

\begin{theorem} (\textbf{Learning rates for group lasso regularization})
\label{cor:gp}

Suppose $(X_t)_{t=1}^T$ follows the SEPP framework of 
  \eqref{eq:model} and $A^\ast$ is estimated using column group lasso regularization.  
 
\begin{enumerate}[label=(\alph*)]  
 \item If $(X_t)_{t=1}^T$ is generated according to the ARMA$(1,1)$ model in  \eqref{eq:arma} then
 $$\|\widehat{A}-A^\ast\|_F^2 \leq  \frac{ C}{R_{\min}^2 \min(\frac{1}{2}R_{\min},\kappa)^2} \frac{s_GM \log^4(MT)}{T}$$ with probability at least $1-\frac{1}{(MT)^c}$ for $T,M$ satisfying $$T \geq 128U^4s_G^2\frac{\log M}{\min(\frac{1}{2}R_{\min},\kappa)^2}.$$ for constants $C,c >0$ which are independent of $M,T$ and $s_G$.
\item If $(X_t)_{t=1}^T$ is generated according to the AR$(2)$ model in  \eqref{eq:ar2} then
 $$\|\widehat{A}-A^\ast\|_F^2 \leq  \frac{ C}{R_{\min}^2 r_\rho^2} \frac{s_GM \log^4(MT)}{T}$$ with probability at least $1-\frac{1}{(MT)^c}$ for $T,M$ satisfying $$T \geq 128U^4s_G^2\frac{\log M}{r_\rho^2}.$$
\end{enumerate}

\end{theorem}

\paragraph{Example 3: Low-rank Regularization}
Estimation of high-dimensional but low-rank matrices is a widely studied problem with numerous applications \cite{rnk, rechtlemma, rnk2, rnk3, rnk4}.  Low-rank models can be seen as a generalization of sparse models, where the matrix is sparse in an unknown basis.  A standard technique to estimate a low-rank matrix is to take a convex relaxation of an $l_0$ penalty on the singular values \cite{rnk4}: the nuclear norm penalty $$\|A\|_{\ast}=\sum_{i=1}^M \sigma_i(A),$$ where $\sigma_i(A)$ denotes the $i$th singular value of $A$.  The dual to the nuclear norm is the operator norm $$\|A\|_{op}=\sup_{\|x\|_2=1} \|Ax\|_2.$$
As discussed in \cite{Neg10}, the nuclear norm is decomposable with respect to the subspaces  
 $$\mathcal{W}=\{A \in \mathbb{R}^{M \times MK} : \text{row}(A) \subseteq \text{row}(A^\ast) \text{ and } \text{col}(A) \subseteq \text{col}(A^\ast) \}$$ and $$\overline{\cal W}^\perp=\{A \in \mathbb{R}^{M \times MK} : \text{row}(A) \subseteq \text{row}(A^\ast)^\perp \text{ and } \text{col}(A) \subseteq \text{col}(A^\ast)^\perp \},$$ where $\text{row}(A)$ and $\text{col}(A)$ denote the row and column spaces of $A$ respectively.
Unlike the previous two examples, here $\cal W \not = \overline{\cal W}$. 

In this low-rank setup, there is no limitation on the number of nodes which can influence a given node.  This introduces challenges in establishing Assumption \ref{A4}, which guarantees near row sparsity of the error.  In order to get around this, we impose a technical assumption on $\|A^\ast\|_{2,1}$ in Lemma \ref{Thm:Rank}.  An area of interest for future work is to examine whether our estimation procedure is flawed when one node can have many nodes influence it, or whether the need for this assumption is an artifact of our analysis.

\begin{lemma}
\label{Thm:Rank}

Suppose $(X_t)_{t=1}^T$ is generated according to  ~\eqref{eq:model} with rank$(A^\ast)=r$ and $\|A^\ast\|_{2,1}^2=D\sqrt{M}$ for a universal constant $D$.  Further, assume $A^\ast$ is estimated according to \eqref{eq:estimator} over the ball $\{A: \|A\|_{2,1}^2 \leq D\sqrt{M}\}$ using nuclear norm regularization.  Then 
\begin{enumerate}[label=(\alph*)]  
\item Assumption \ref{A3} is satisfied with $$\Psi(\overline{\cal W})=\sqrt{2r}.$$
\item Assumption \ref{A4} is satisfied with $$\mu_{\ast}=2D\sqrt{M}.$$
\item Assumption \ref{A5} is satisfied with  $$\|\frac{1}{T} \sum_{t=1}^T \epsilon_t g(\mathcal{X}_t)^\top\|_{op} \leq \log^4(MT) \sqrt{\frac{M}{T}}$$ with probability at least $1-\frac{1}{(MT)^c}$.
\end{enumerate}
\end{lemma}

\paragraph{Proof Overview}
\begin{itemize}
    \item The main challenge in Lemma \ref{Thm:Rank} comes in the proof of the deviation bound condition, which depends on the concentration properties of vector-valued martingales. The concentration properties of 2-smooth norms was studied in a number of works, including \cite{smooththm,smooththm2}. We leverage recent work in \cite{regularthm} which extends the concentration results for 2-smooth norms to operator norms.
    
    \item Assumption \ref{A4} follows from assuming $\|A^\ast\|_{2,1}$ and $ \|\widehat{A}\|_{2,1}$ are on the order of $\sqrt{M}$.  Without this assumption, we could potentially have $\mu_\ast = O(M)$.  This would give us a final bound in Corollary \ref{cor:rnk} which is only applicable when $T \geq M^2$, so this technical assumption is crucial in constructing a meaningful bound.  
    \item The subspace compatibility constant in Assumption \ref{A3} was shown in \cite{rechtlemma}.  In the sparsity case, this condition is trivial because $\cal S=\overline{\cal S}$ and thus $\triangle_{\overline{\cal S}}$ is known to lie in a subspace where every element is $s$-sparse.  The condition is more subtle in the nuclear norm regularization case because $\cal W =\overline{\cal W}$ if and only if $A^\ast$ is symmetric.  We do not assume symmetry of $A^\ast$ so $\triangle_{\overline{\cal W}}$ need not lie in the subspace $\cal W$ where each element has rank at most $r$.  However, \cite{rechtlemma} shows that $\overline{\cal W}$ only contains matrices of rank at most $2r$. 
\end{itemize}

Combining Lemma \ref{Thm:Rank}, Theorem \ref{Thm:bound} and the restricted eigenvalues results gives the following Theorem.

\begin{theorem} (\textbf{Learning rates for nuclear norm regularization})
\label{cor:rnk}

Suppose $(X_t)_{t=1}^T$ follows the SEPP framework of 
  \eqref{eq:model} and $A^\ast$ is estimated using nuclear norm regularization over the ball $\{A: \|A\|_{2,1}^2 \leq D\sqrt{M}\}$.
 
\begin{enumerate}[label=(\alph*)]  

\item If $(X_t)_{t=1}^T$ is generated according to the ARMA$(1,1)$ model in  \eqref{eq:arma} then
 $$\|\widehat{A}-A^\ast\|_F^2 \leq  \frac{ C
   }{R_{\min}^2 \min(\frac{1}{2}R_{\min},\kappa)^2} \frac{rM \log^8(MT)}{T}$$ with probability at least $1-\frac{1}{(MT)^c}$ for $T,M$ satisfying $$T \geq 128U^4M\frac{\log M}{\min(\frac{1}{2}R_{\min},\kappa)^2}$$ for constants $C,c >0$ which are independent of $M,T$ and $r$.

\item If $(X_t)_{t=1}^T$ is generated according to the AR$(2)$ model in  \eqref{eq:ar2} then
 $$\|\widehat{A}-A^\ast\|_F^2 \leq  \frac{ C
   }{R_{\min}^2 r_\rho^2} \frac{rM \log^8(MT)}{T}$$ with probability at least $1-\frac{1}{(MT)^c}$ for $T,M$ satisfying $$T \geq 128U^4M\frac{\log M}{r_\rho^2}.$$

\end{enumerate}

\end{theorem}
Once again the mean-squared error bound $\frac{rM \log^8(MT)}{T}$ matches the minimax optimal rate for independent design~\cite{Neg12} up to log factors.

\section{Numerical experiments}

We validate our methodology and theory using a simulation study and real data examples. The focus of the simulation study is to confirm that the rates in mean-squared error in terms of $s$, $r$, $T$ and $\kappa$ scale as the theory predicts.  We generate data according to the ARMA$(1,1)$ model from  \ref{eq:arma}. 

Our focus with real data experiments is  to demonstrate that the models we analyze are sufficiently complex to capture real-world phenomena and enhance prediction performance relative to naive models. Others have successfully used more complex, difficult to analyze models (\cf \cite{linderman2014discovering, ertekin2015reactive}) which are similar in spirit to those analyzed here. Our claim is not that our approach leads to uniformly better empirical performance than previous methods, but rather that our models capture essential elements of all these approaches and hence our theoretical work provides insights into a variety of approaches.

Our first real data example shows that our model and estimation procedure determines interactions among shooting events across different communities of Chicago that obeys sensible spatial structure (even though the algorithm does not use any spatial information).  Our second real data example looks at neuron firing patterns in the brain of a rat and shows that our model can differentiate between the firing patterns during a sleep period and the patterns during a wake period.  Finally, we examine a data set consisting of articles posted by different news websites and we try to determine influences between the sites using a variety of different regularization techniques.  We implement the estimation procedure in  \eqref{eq:estimator} using the SpaRSA algorithm from \cite{sparsa}.

\subsection{Simulation study} 

We generate data according to  \eqref{eq:arma} with $\nu=0$, $M=50$, $\tilde U=6$, $\alpha=.25$ and varying values for $T$ and $s$.  Recall that $\nu$ controls the background rate, $M$ is the number of nodes, $\tilde U$ is the clipping threshold, $\alpha$ is the memory of the process in \eqref{eq:arma}, $T$ is the number of time steps observed, and $s$ is the number of edges in the network.

Each time we generate a matrix, we randomly select $s$ entries to be nonzero, and assign each value uniformly in $[-.7,.3]$.  The sparsity ranges between $10$ and $50$.  With these parameters, our process is usually stable on its own, and only occasionally relies on the clipping function to dampen the observations.  Even at $s=50$, only around $1\%$ of the observations exceed $6$, and the clipping percentage is even lower for smaller $s$. For each choice of $s,T$, we run $100$ trials with $\lambda=.1/\sqrt{T}$.  In the $i^{\rm}$ trial we form a ground truth matrix $A^\ast_i$, compute $\hat A_i$, and measure the mean squared error (MSE) as $\|A^\ast_i - \hat A_i\|_F^2$.

\begin{figure}[ht]
\centering
\subfloat[MSE vs $T$]{\includegraphics[width=.5\linewidth, height=4cm]{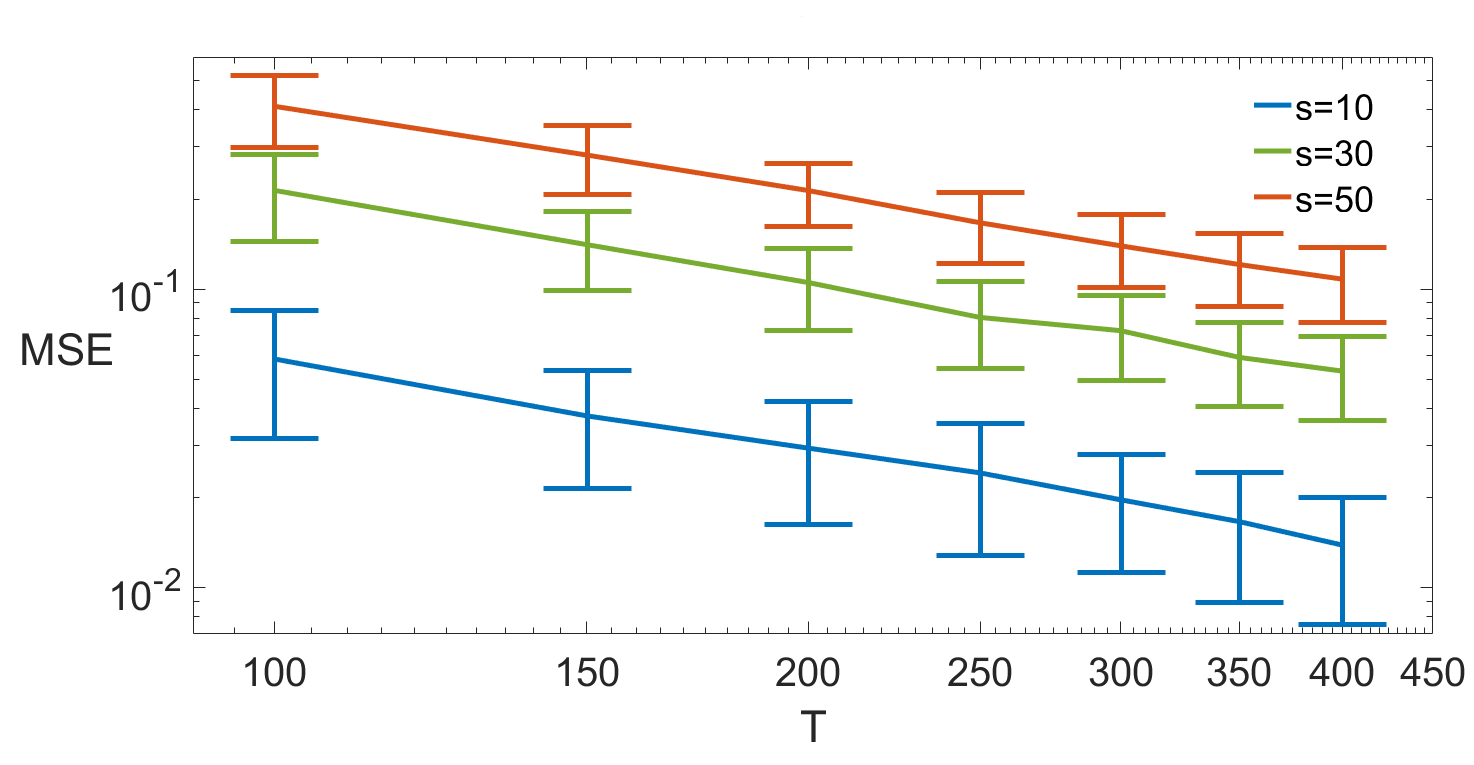}}~
\subfloat[MSE vs $s$]{\includegraphics[width=.5\linewidth, height=4cm]{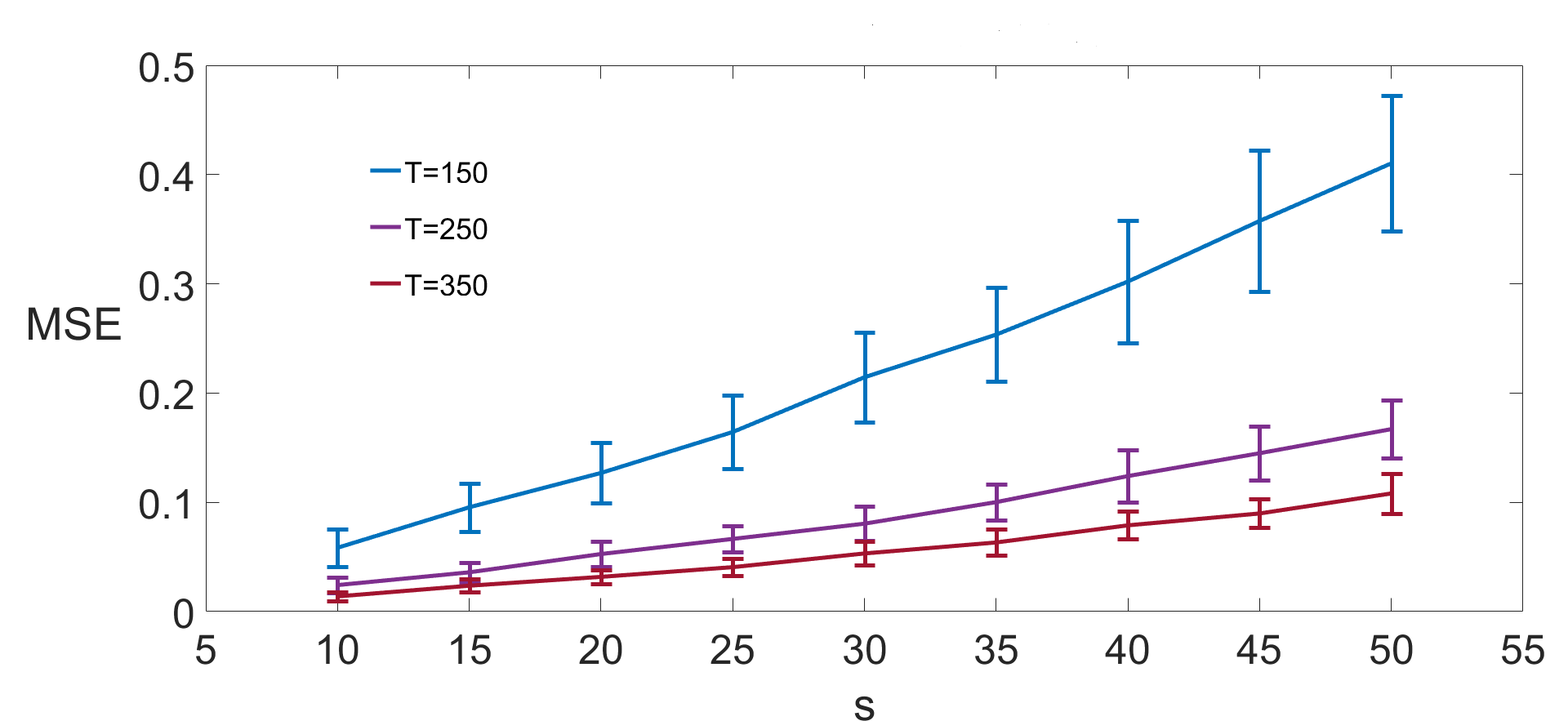}
}
\caption{(a) shows MSE vs $T$ for varying values of $s$, while (b) shows MSE
  vs $s$ for varying values of $T$.  Plots agree with theory which
  suggests that MSE scales linearly in $s$ and $\frac{1}{T}$.  Median
  of $100$ trials are shown, and error bars denote the standard
  deviation.}
\label{fig:sims}
\end{figure}

In Figure~\ref{fig:sims}(a) , we plot $\mbox{MSE} $ vs $T$ for several
different values of $s$, and in Figure~\ref{fig:sims}(b) we plot
$\mbox{MSE}$ vs $s$ for several values of $T$.  The plots agree with our theory, which suggests
that the error scales linearly in $s$ and $\frac{1}{T}$.

\begin{figure}[ht]
\centering
\subfloat[MSE vs $T$]{\includegraphics[width=.5\linewidth, height=4cm]{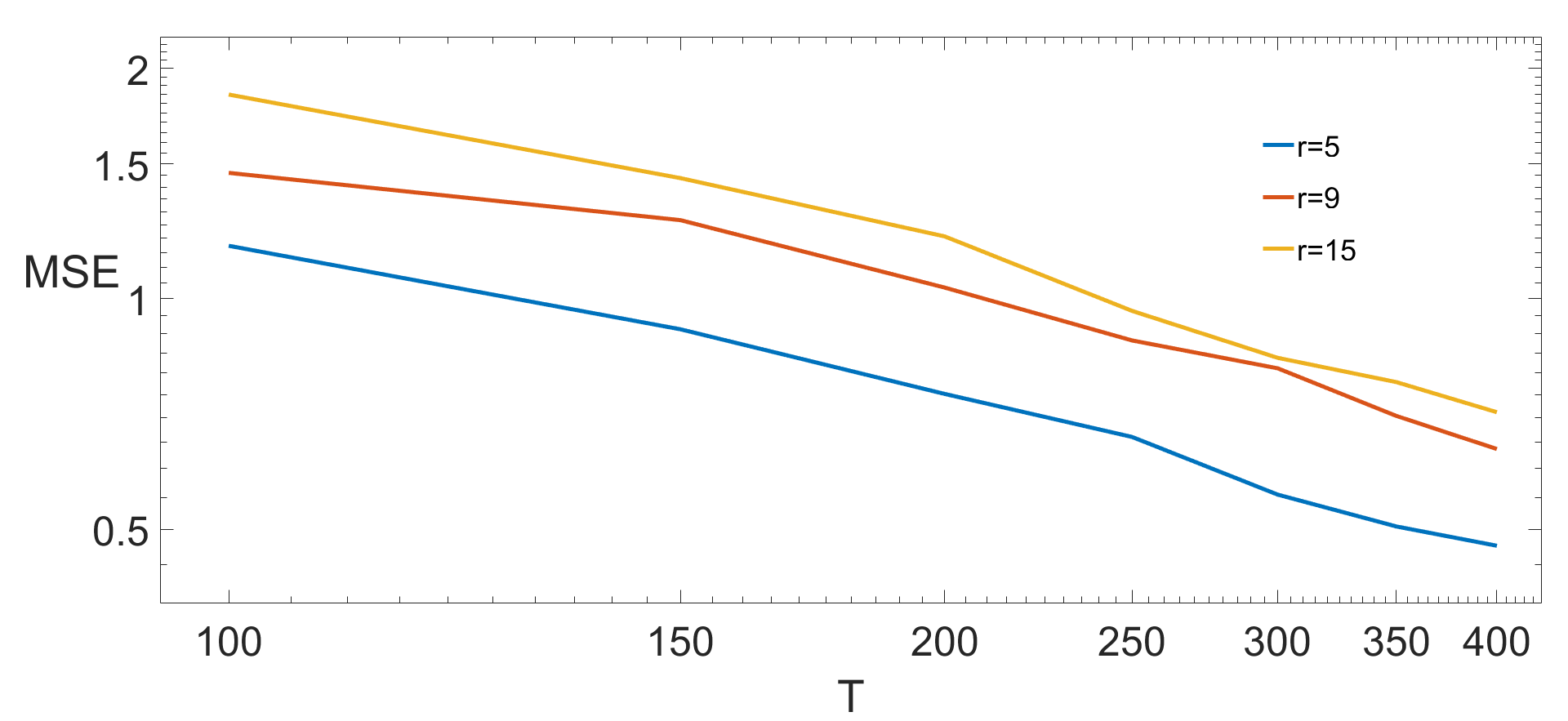}}~
\subfloat[MSE vs $r$]{\includegraphics[width=.5\linewidth, height=4cm]{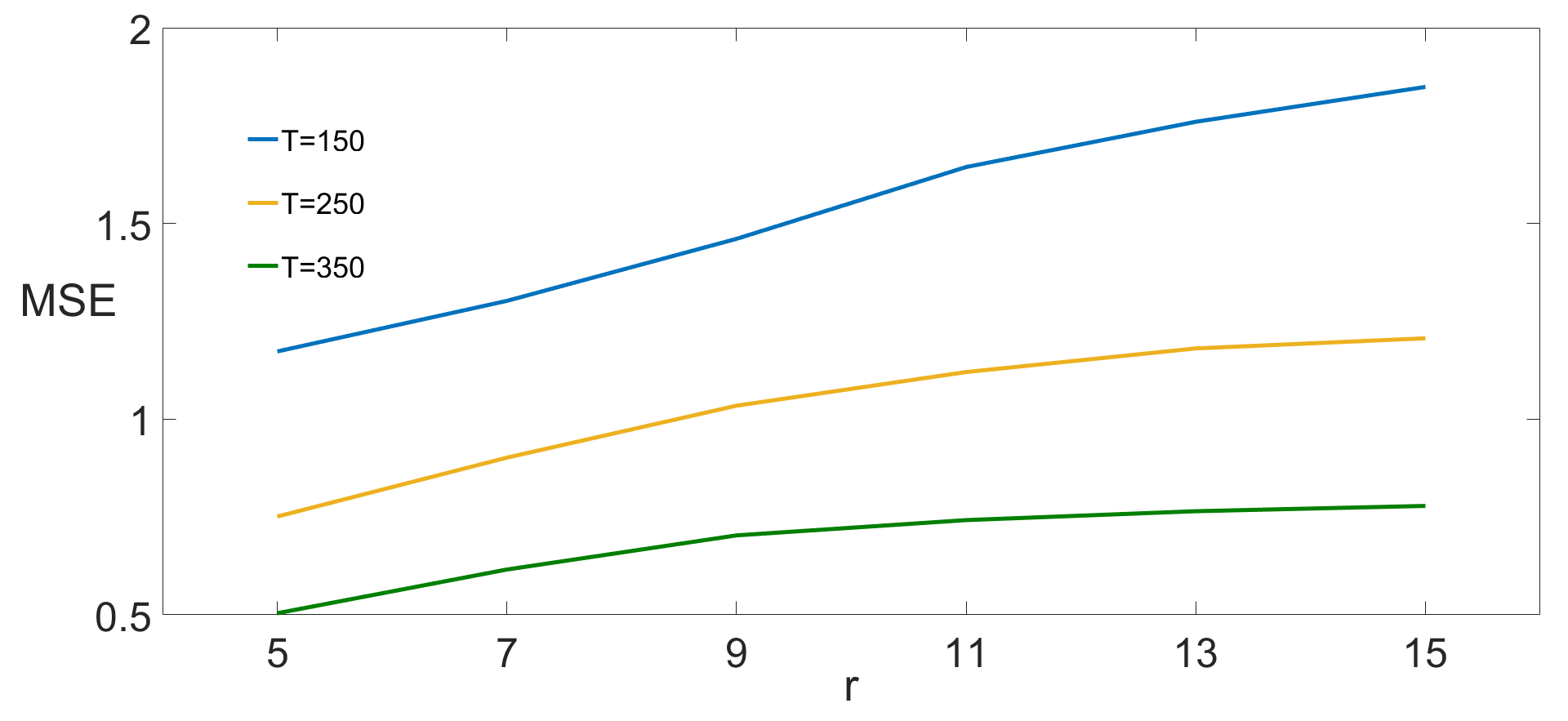}
}
\caption{(a) shows MSE vs $T$ for varying values of $r$, while (b) shows MSE
  vs $r$ for varying values of $T$.  Plots agree with theory which
  suggests that MSE scales linearly in $r$ and $\frac{1}{T}$.  Median
  of $100$ trials are shown.}
\label{fig:sims2}
\end{figure}

Next, instead of assuming that $A^\ast$ only has $s$ non-zero entries, we assume that $A^\ast$ has rank $r$, and measure MSE as a function of $r$.  We hold the remaining parameters the same.  To generate a rank $r$ matrix $A^\ast$, we randomly generate a $M \times r$ matrix and multiply it by a randomly generated $r \times M$ matrix where the entries of both matrices are uniformly drawn from $[-.7,.3]$.  We then normalize the resulting matrix so that $a_{\max}$ is approximately $.3$.  For all choices of rank considered, less than $5\%$ of observations are clipped.  We set $\lambda=.1\sqrt{\frac{M}{T}}$ as guided by our theory, run $100$ trials for each $(r,T)$ pair, and plot the median in Figure \ref{fig:sims2}.  The simulations agree with Lemma \ref{Thm:Rank} which suggests that the MSE should scale linearly in $r$ and $\frac{1}{T}$.  

We now examine the relationship between $a_{\max}$, our theoretical MSE, and simulated MSE.  For the remainder of the section fix $\tilde U=6$ and $\alpha=0$ and $T=400$.  Recall from Figure~\ref{fig:heatmap} that there is a stark phase transition where $\kappa$ goes from reasonably large to minuscule.  For $\tilde{U}=6$, this transition occurs between $a_{\max}=.3$ and $a_{\max}=.4$.  Our MSE bound scales with $\kappa^{-2}$,  so this phase transition controls where our bound is reasonably small. 

To get a sense of how tight the bound is, we consider two different methods to generate a $50 \times 50$ matrix $A^\ast$.  The first is a block design, where $A^\ast$ is zero outside of five $10 \times 10$ blocks on the diagonal.  Within the blocks, each row has five nonzero entries picked at random with values equal to $\frac{a_{\max}}{5}$.  Matrices with this structure have strong feedback loops, where large observations from one node stimulate other nodes which are likely to feedback to the original node.  In other words, with this block design method it is likely that many observations will actually be drawn close to the maximum possible rate $R_{\max}$, so we expect the MSE to align closely with our theoretical bound.  We estimate $A^\ast$ using $l_1$ regularization, for varying values of $a_{\max}$.  

As a second method, we consider a low-rank design and estimate $A^\ast$ using nuclear norm regularization.  We choose the first two rows of $A^\ast$ to be orthogonal, both with row sums equal to $a_{\max}$.  We then let each remaining row be a random convex combination of the first two rows.  In this case, feedback loops are less of an issue; if one node has a large observation, the nodes it stimulates are less likely to have strong connections feeding back to the original node.  Since this design method is not particularly stimulatory, we expect that most observations will not be drawn close to the maximum rate $R_{\max}$.  Our theoretical bound is potentially loose in this case for the following reason. Our bound on $\kappa$ which captures the amount of clipping is worst case based on the size of the coefficients of $A^\ast$, but does not take into account the structure of $A^\ast$.  If the coefficients of $A^\ast$ are large but are structured such that there aren't a lot of feedback loops then our bound on $\kappa$ will be loose.

\begin{figure}

	\centering
	\includegraphics[width=.8\linewidth]{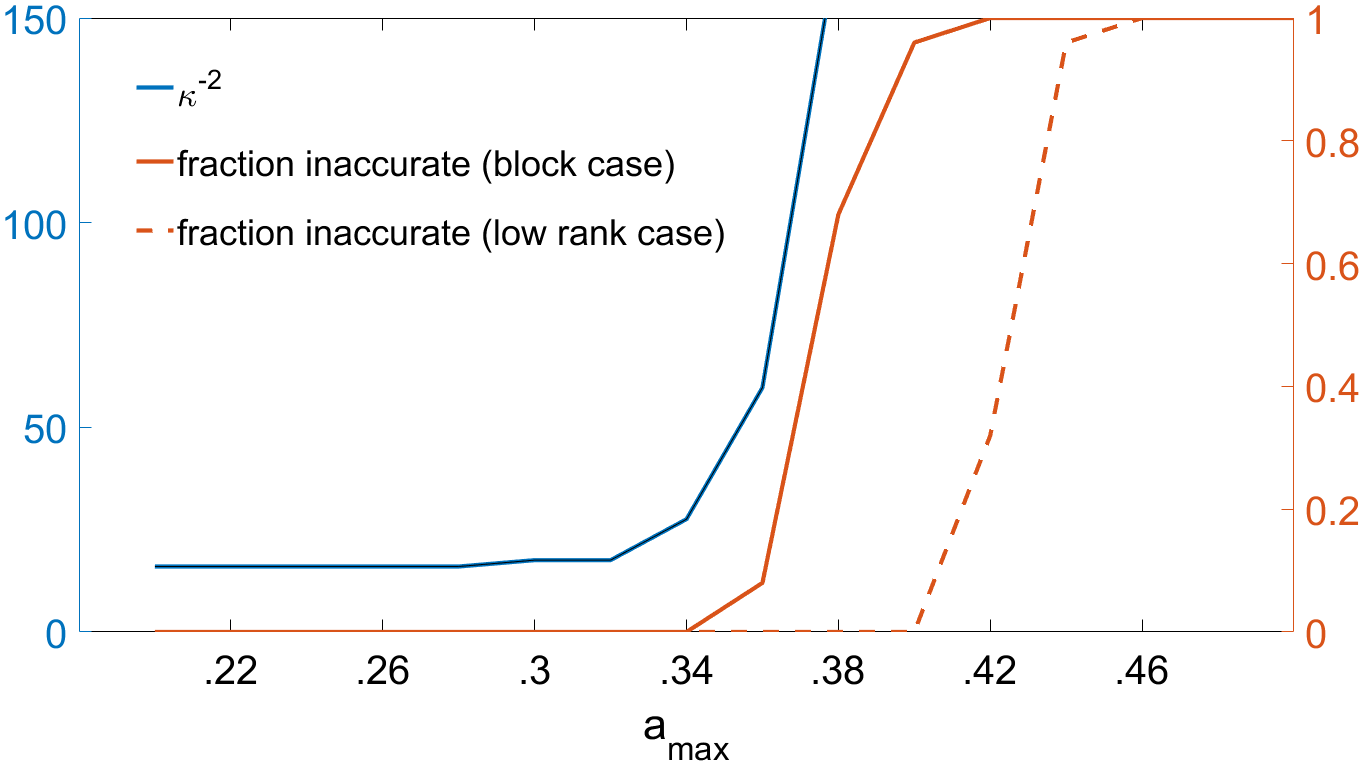}
	\caption{In blue, $\kappa^{-2}$ as a function of $a_{\max}$.  In orange, the fraction of 50 trials which have MSE above 1 for a block matrix design strategy and a low-rank matrix design strategy.  Our theoretical bound on the MSE scales with $\kappa^{-2}$, suggesting stark phase transitions in the MSE which are confirmed in the simulated results.  The block matrices correspond to more stimulatory networks than the low-rank ones, and therefore align more closely with our worst case bounds.  }
\label{fig:mse_vs_kappa}
\end{figure}

We randomly generate $50$ different $A^\ast$ over various $a_{\max}$ for both design choices and then evaluate their efficiency by plotting the fraction of trials for which the MSE is above one.  The results are shown in Figure ~\ref{fig:mse_vs_kappa}.  The simulated results also exhibit strong phase transitions, with the fraction of accurate trials shifting from one to zero with small changes in $a_{\max}$.  This suggests that our theoretical results capture a real phenomenon of our model.  In the block case, the phase transition occurs almost exactly where predicted by the MSE, whereas in the low-rank case there is a small lag in the phase transition.  In other words, while our theoretical results are fairly tight for very stimulatory network structures, there appears to be some flexibility for networks with weaker feedback loops.

\begin{figure}

	\centering
	\includegraphics[width=.8\linewidth]{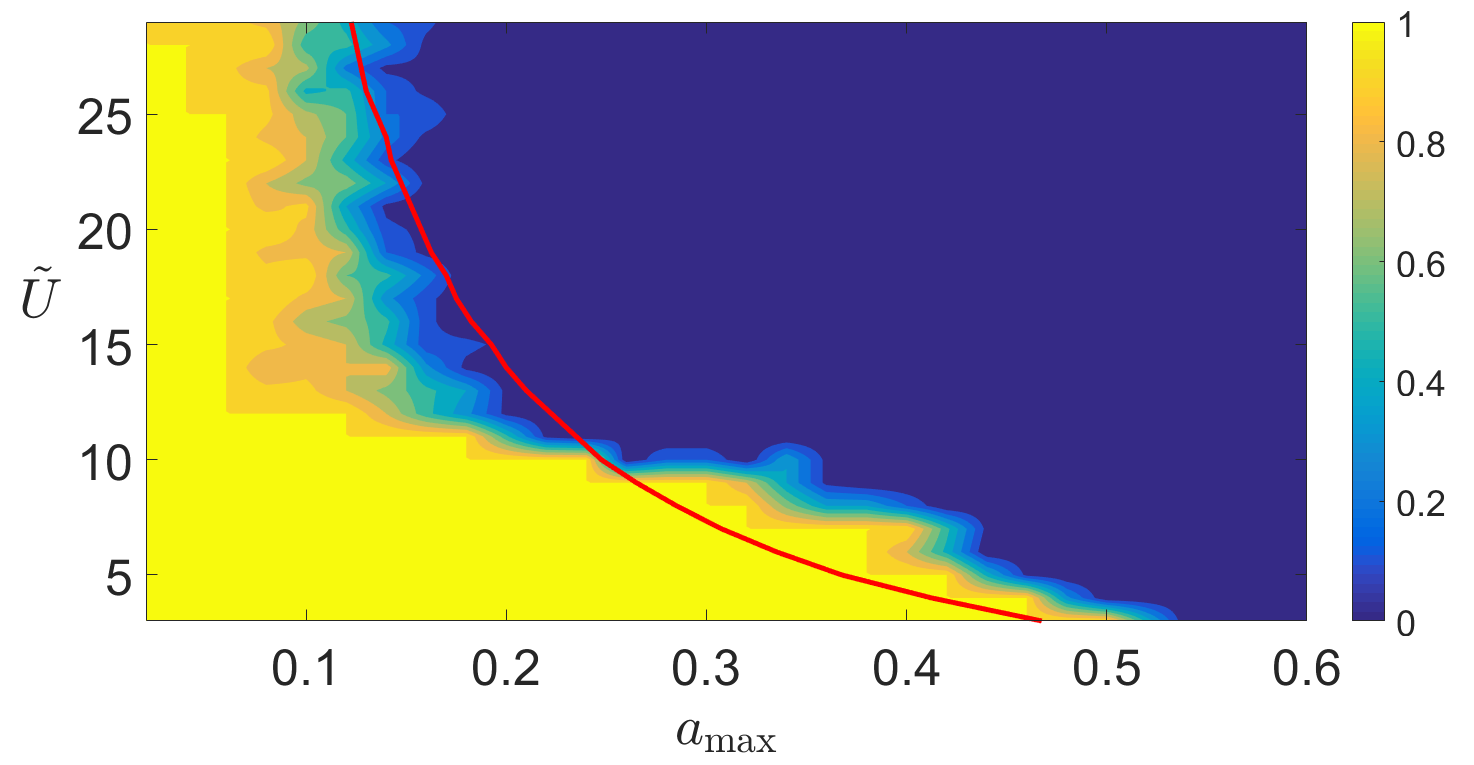}
	\caption{Heat map displaying the fraction of trials for which the MSE is below one for different $(a_{\max},\tilde U)$ pairs.  Red line shows $\kappa=.01$ contour.  Above this line our theory predicts difficulty in learning, agreeing with the heat map which shows inaccurate recovery of $A^\ast$ for $(a_{\max},\tilde U)$ pairs above the line.  For each $(a_{\max},\tilde U)$ pair, we run $20$ trials with a block matrix design.}
\label{fig:mse_vs_kappa_sim}
\end{figure}

Finally, we consider the block matrix design under a wide range of $\tilde U$.  We consider $\tilde U$ between $3$ and $30$, and $a_{\max}$ between $0$ and $.6$ in increments of $.02$.  For each $(a_{\max},\tilde U)$ pair, we generate 20 matrices according to the block matrix design strategy outlined above and estimate $\widehat{A}$ via sparsity regularization. In Figure \ref{fig:mse_vs_kappa_sim} we plot a heat map displaying the fraction of trials for with the MSE is below one.  The red line shows the $\kappa=.01$ contour, so Lemma \ref{Thm:Sparsity} suggests our model will be hard to learn above the boundary line.  Figure \ref{fig:mse_vs_kappa_sim} generally resembles the heat map displaying values of $\kappa$ in Figure \ref{fig:heatmap}, suggesting that the role of $\kappa$ in our theory reflects a true phenomenon that when $a_{\max}$ is sufficiently large and clipping is frequent, then the model becomes difficult to learn.  

\subsection{Real Data Example -- Chicago Crime}

 A number of studies have used various self-exciting point processes to predict crime, including \cite{BertozziHawkes, mohler2012chi, mohler2014hotspot}.  We test our model on a data set \cite{citychicago} consisting of burglaries in Chicago since 2004, broken down by the $M=77$ community areas in the city.  In \cite{latentnet}, the authors fit self-exciting processes to the Chicago homicide data broken down by community area and performed clustering on the areas as we do below.  We estimate the network based on the data from January 2004 to August 2010 and test it  on the data from September 2010 to March 2017. To test results, we compare the log-likelihood of events using our learned matrix on the test set data, with that for a constant Poisson process.  This gives approximately $600$ time periods for both our training and test sets.  We set $\lambda=.1/\sqrt{T}$ using our theory as a guide.  We show results for a half-day time discretization period, with $\tilde U=7$ and $\alpha=.2$.  
 
 The test set log-likelihood of our learned matrix shows an improvement over the test set log-likelihood for a learned constant process, where $\lambda_{t,m} = \overline \lambda_m$ for all $t$ ($-6.62 \times 10^5$ compared to $-1.09 \times 10^{6}$).

\begin{figure}[ht]
	\centering
	\includegraphics[width=.7\linewidth]{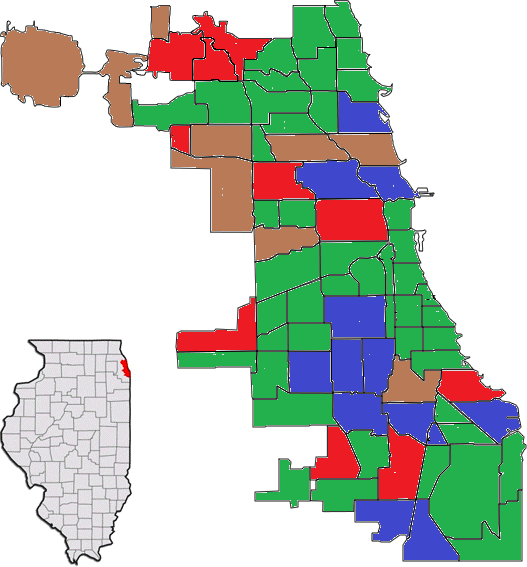}	
	\caption{Clusters learned from crime data with a half-day time discretization.
          The clusters are overlaid on a map of community areas in
          Chicago.  The data contained no geospatial information, but clusters show geographical patterns. }
	\label{fig:chicago}
\end{figure}

To examine the structure of our learned matrix, we treat the positive coefficients of the matrix as edges of the adjacency matrix of a graph.  We then perform spectral clustering with four clusters. 
The results are shown in Figure~\ref{fig:chicago}, with colors indicating cluster membership.  We note that  our data contains no information about the geospatial location of the areas aside from index (not location) of the community area. However, there are clear geographic patterns in the clusters, providing some validation to the estimated influences between communities.

Finally, we test whether modeling these crime patterns as a multi-dimensional point processes leads to stronger results than modeling the patterns as a collection of independent univariate point processes, where $$\log \lambda_{t+1,m} =\nu_m + a^\ast_{m} \min(X_{t,m},\tilde U) + \alpha\log(\lambda_{t,m}).$$ Specifically, we compare finding $\widehat{A}$ as in \eqref{eq:estimator} with choosing $\widehat{A}$ to be the solution to the optimization problem in \eqref{eq:estimator} over the set of all diagonal matrices.  We then perform the log-likelihood analysis described above for $\alpha$ varying from $0$ to $.6$ in increments of $.2$ and for the time-discretization period varying from half a day to three days in increments of half a day.  The results are shown in Table \ref{table:one}.  Note that the multivariate model outperforms the univariate one whenever the discretization period is at least a full day but does worse for the half day discretization period. 

\begin{table}[ht]
\caption{Difference between Log-Likelihood of multivariate process and univariate process} 
\centering 
\begin{tabular}{c c c c} 
\hline\hline 
Discretization & $\alpha=0$ &$\alpha=.2$ & $\alpha=.4$  \\ [0.5ex] 
\hline 
.5 days & $-5.4 \times 10^4$ & $-6.5\times 10^4$ & $-6.6 \times 10^4$ \\ 
1 day & $1.2 \times 10^5$ & $7.6 \times 10^4$ & $6.2 \times 10^4$  \\
1.5 days & $1.8 \times 10^5$ & $1.1\times 10^5$ & $1.4 \times 10^5$ \\
2 days & $1.8 \times 10^5$ & $1.2 \times 10^5$ & $2.3 \times 10^5$  \\
2.5 days & $2.1 \times 10^5$ & $1.1 \times 10^5$ & $2.7 \times 10^5$ \\
3 days & $1.9 \times 10^5$ & $1.1\times 10^5$ & $2.8 \times 10^5$ \\
 [1ex] 
\hline 
\end{tabular}
\label{table:one} 
\end{table}

\subsection{Real Data Example -- Spike Train Data}
\begin{figure}[ht]
	\centering
	\subfloat[$A_{\text{wake}}$]{\includegraphics[width=.5\linewidth]{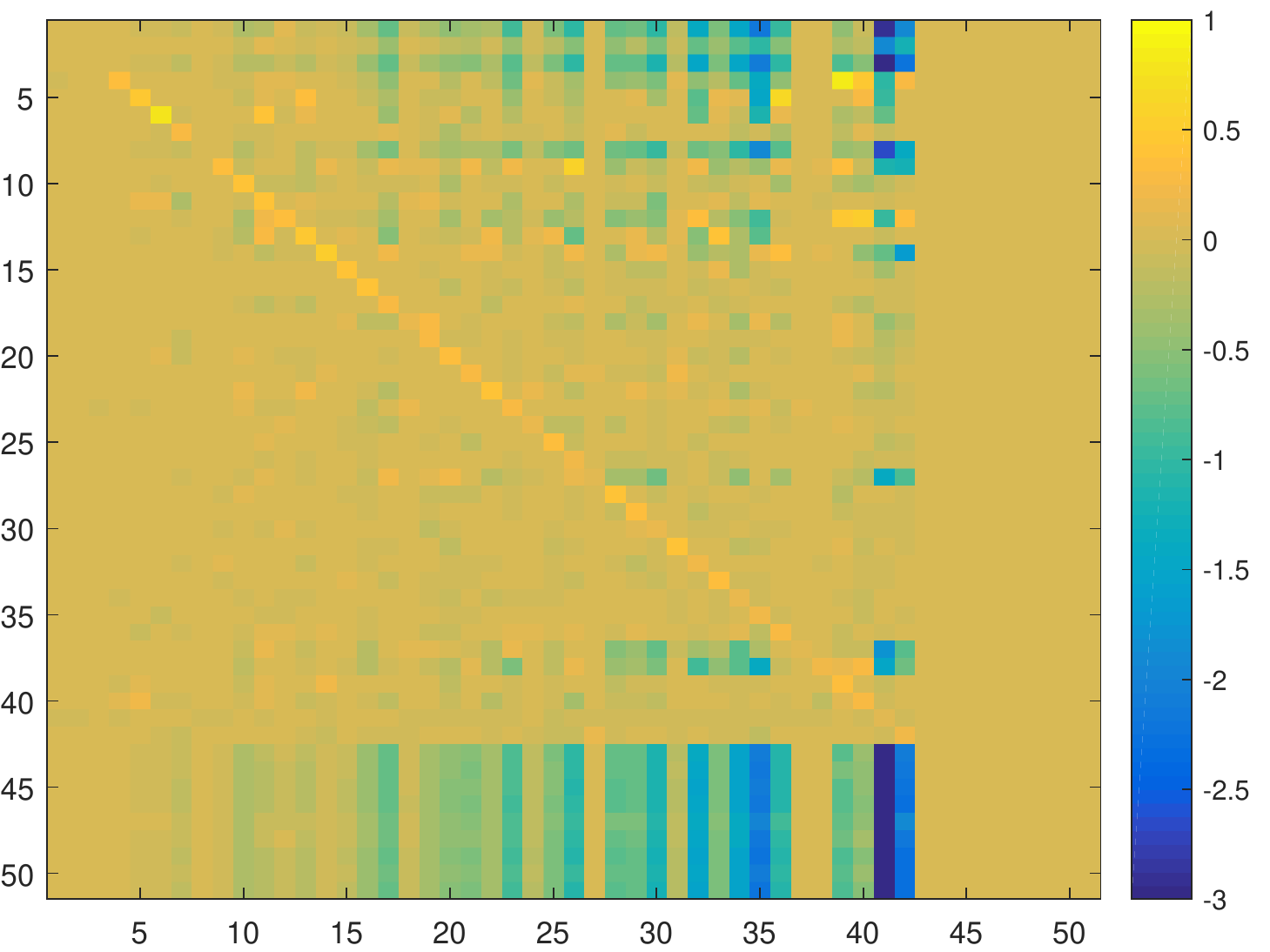}}~
\subfloat[$A_{\text{sleep}}$]{\includegraphics[width=.5\linewidth]{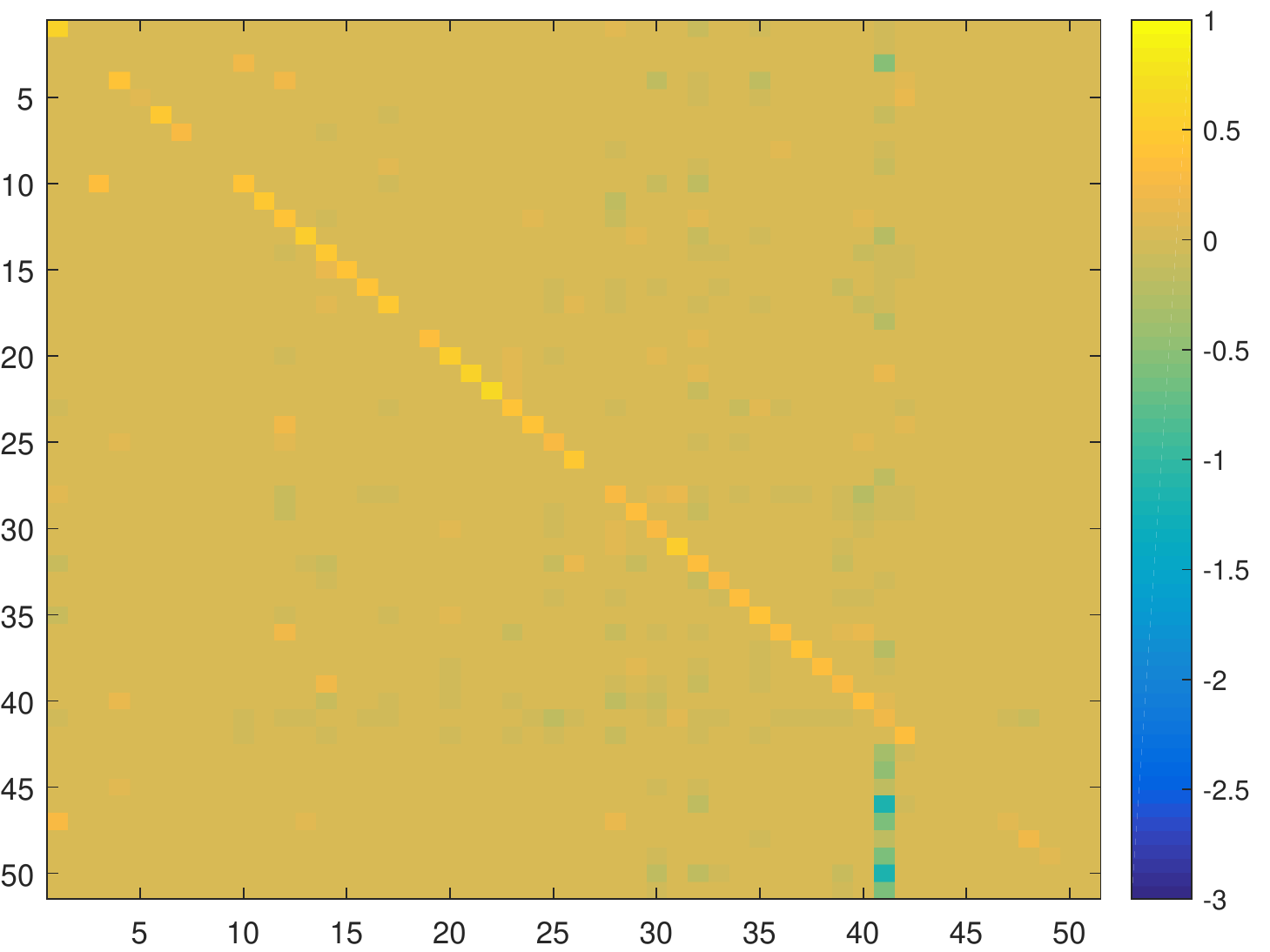}
}
	\caption{(a) shows $A_{\text{wake}}$ matrix charting estimated relationships between neurons in a rat's brain during a wake period, while (b) shows $A_{\text{sleep}}$ matrix charting estimated relationships between neurons in a rat's brain during a sleep period.  }
	\label{fig:neuro}
\end{figure}

SEPPs have been widely used in neuroscience to describe neuron spike train data \cite{lindermanneuro, ppglm3, ppglm, ppglm2}.  In this section, we analyze a multi-neuron spike train dataset from \cite{dataset, datasetpaper}. The dataset consists of spike trains recorded from $51$ neurons in the brain of a rat.  The recordings were divided into a wake period and a sleep period.  Using the data from the first half of the wake period, we learn a matrix $A_{\text{wake}}$ using equation \eqref{eq:arma}, a $100$ms discretization period, $\alpha=.7$ and $\tilde U=5$.  We then follow the same process to learn $A_{\text{sleep}}$.  We get a sense of the structure of the matrices $A_{\text{wake}}$ and $A_{\text{sleep}}$ in Figure \ref{fig:neuro}.  We note that connections are much stronger during the wake period, during which there is more frequent neural firing. 

In  previous work \cite{kasspaper} the authors use a similar SEPP to analyze neural spikes and discuss the significance of the time discretization period in more depth.  In particular, they conclude that while models at this discretization length may have strong predictive power, the discretization period is sufficiently large that the connections learned are not direct physical effects.  In other words, if the connection between neuron A and neuron B is negative, this suggests that neuron B is less likely to fire in a $100$ms interval after neuron A fires.  However, there could be a complex chain of interactions causing this effect, and it does not mean there is a direct physical connection between neuron A and neuron B.

To validate $A_{\text{wake}}$, we compute the log-likelihood of events for the second half of the wake period using both $A_{\text{wake}}$ and $A_{\text{sleep}}$ as the ground truth matrix.  
We find $\log p(X_{\text{wake}}|\hat A_{\text{wake}})=-6.6 \times 10^4$ while $\log p(X_{\text{sleep}}|\hat A_{\text{wake}})=-7.4 \times 10^4$. 

Following the same process for $A_{\text{sleep}}$, we find that $\log p(X_{\text{sleep}}|\hat A_{\text{sleep}})=-2.34 \times 10^5$ while $\log p(X_{\text{wake}}|\hat A_{\text{sleep}})=-3.04 \times 10^5$.  This suggests that our model is capable of differentiating firing patterns in different sleep states.  The log-likelihood of events for a constant process was orders of magnitude smaller in both cases. 

\subsection{Real Data Example -- Memetracker Data}As a final example, we consider a data set \cite{meme} which consists of metadata for a collection news articles and blog posts.  We only consider the time and website from which each post occurs but omit all other data such as the content of the post and other websites to which the post links.  Further, we consider only articles posted by $198$ popular news sites from \url{http://www.memetracker.org/lag.html}.  Low-rank models have been applied in social network settings in a number of different works \cite{zhouZhaSongHawkes, community, link};  in particular, the work \cite{zhouZhaSongHawkes} proposes low-rank regularization of a point process model on this same data set.

\begin{figure}[ht]
	\centering
	\includegraphics[width=.7\linewidth]{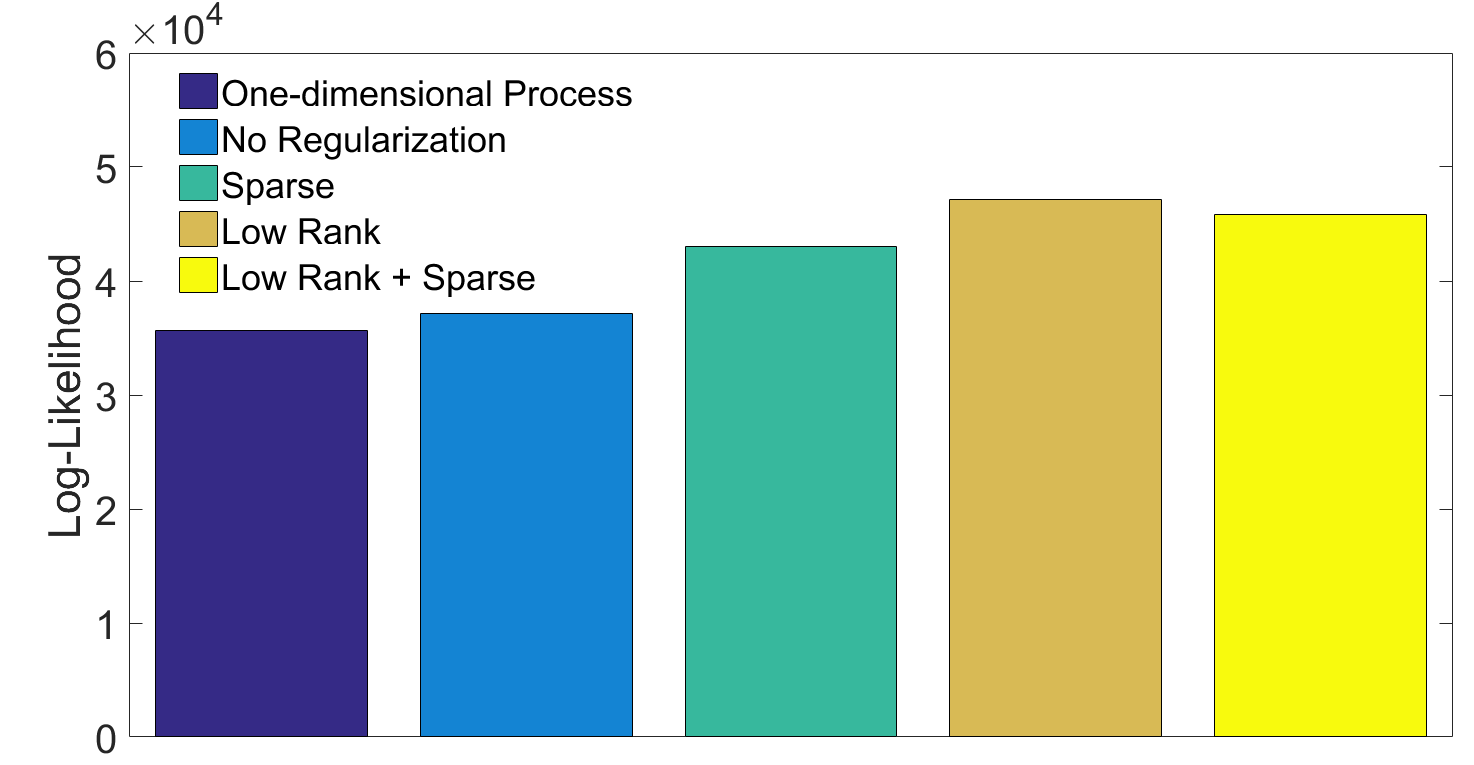}	
	\caption{Log-likelihood of events on test set for matricies learned using memetracker data set under a variety of structure assumptions.  Data set consists of timestamps for articles posted by $198$ popular news websites and blogs during October 2008.}
	\label{fig:memetracker}
\end{figure}

To test the model, we collect all articles posted by $198$ popular new websites during October 2008.  Using a one hour discretization period, we divide the month into a training set and a test set, giving $T=500$ training periods and $500$ test periods.  We train our model using the following regularization techniques.  We perform the $l_1$ regularization and nuclear norm regularization schemes described in Section \ref{sec:regularization}, as well as a low-rank plus sparse model where we use the regularizer $$\|\cdot\|_{\cal R}=\|\cdot\|_{1,1}+\|\cdot\|_{\ast}.$$  This last model is optimized using alternating descent.  Finally, we learn a multi-dimensional model with no regularization, where we simply use the negative log-likelihood as our loss function, and a one-dimensional model where all interactions between different nodes are set to zero.  The results are shown in Figure \ref{fig:memetracker}.  The low-rank model performs best, followed by the low-rank plus sparse model, suggesting that the interactions between websites exhibit some low-rank behavior.

\section{Connections to Hawkes Process}

In this section, we observe that the model in  \eqref{eq:model} can be seen as a discretized version of the multivariate Hawkes process, in which there is much long-standing and recent interest (\eg \cite{hawkes1,hawkes2,PointProcesses,clipped_hawkes, zhouZhaSongHawkes, soloHawkesICASSP13, granger}).  By formulating our discrete time model in this manner we aim to highlight the connections between the two classes of models.  There are advantages to analyzing point process models from both the continuous and discrete perspective.  Some advantages of the discrete perspective include:

\begin{enumerate}

\item Real world data comes inherently discretized.  In some cases, \eg social media posts, one might record data accurately up to very fine time windows.  However, in other problems, the data collection process forces a coarse discretization.  For example in \cite{lewis2012self}, which used Hawkes processes to model civilian deaths in Iraq, reliable data was only obtained for the day on which attacks occurred.

\item In many works the authors discuss continuous Hawkes models, but their algorithms work with discretized data for computational efficiency.  Examples include \cite{ait2010modeling,lewis2012self,ppglm}.  This provides additional motivation for our decision to study the ARMA$(1,1)$ model because its continuous version is one of the most widely studied types of Hawkes process.

\end{enumerate}

\subsection{Continuous Hawkes} 
In a multivariate Hawkes process, point process observations are drawn
using an intensity function $\lambda(\mathcal{X}_\tau)$, where $\mathcal{X}_\tau$ is the collection of
all events up to (continuous) time $\tau$. Each event $i$ is associated with two
components: $(\tau_i,m_i)$, where $\tau_i$ is the time of the event
and $m_i$ is the node or channel associated with the event. $N_\tau$
denotes the number of events before time $\tau$.

We model the log-linear Hawkes process intensity at node $m$
as\footnote{$\lambda_m^\cont(\mathcal{X}_\tau)$ would be more precisely written as
  $\lambda_m^\cont(\tau;\mathcal{X}_\tau)$; we let the dependence on $\tau$ be
  understood for simplicity of presentation.}
\begin{align}
\log \lambda_m^\cont(\mathcal{X}_\tau) =& \nu_m + \sum_{i=1}^{N_\tau}
                                h_{m,m_i}(\tau - \tau_i), \label{eq:hawkes_h}
\end{align}
where the ${(c)}$ superscript denotes continuous time. 

Here each function $h_{m,m'}(\tau)$ measures the influence of node
$m'$ on node $m$ after $\tau$ seconds since the event on $m'$. This
model is standard in the point process literature. We write each of
these functions as a linear combination of the basis functions $\phi_1(\tau),\ldots ,\phi_K(\tau)$:
\begin{align}
h_{m,m'}(\tau) = \sum_{k=1}^K a^*_{m,m',k} \phi_k(\tau),\label{eq:hphi}
\end{align}
yielding
\begin{align}
\log \lambda_m^\cont(\tau) =& \nu_m + \sum_{i=1}^{N_{\tau}} \sum_{k=1}^K
a^*_{m,m_i,k}\phi_k(\tau-\tau_i) \nonumber\\
=& \nu_m + \sum_{m' = 1}^M a^*_{m,m'}
   \left[\sum_{\substack{i<N_\tau:\\m_i=m'}} \sum_{k=1}^K
  \phi_k(\tau-\tau_i)\right]\nonumber \\
 =& \nu_m + \sum_{m' = 1}^M \sum_{k=1}^K a^*_{m,m',k} g_{m,k}^\cont(\mathcal{X}_\tau), \label{eq:hawkes}
\end{align}
where
$$g_{m,k}^\cont(\mathcal{X}_\tau) := \sum_{\substack{i<N_\tau:\\m_i=m}} \phi_k(\tau-\tau_i).$$
Vectorizing across nodes and letting
\begin{align*}
\lambda^\cont(\mathcal{X}_\tau) :=&\; [\lambda_1^\cont(\mathcal{X}_\tau),\cdots,\lambda_M^\cont(\mathcal{X}_\tau)]^\top\in \reals^M_+\\
\nu :=&\; [\nu_1,\cdots,\nu_M]^\top \in \reals^M\\
 g^\cont(\mathcal{X}_\tau) :=&\; (g_{m,k}^\cont(\mathcal{X}_\tau))_{m\in\{1,\ldots,M\}, k \in \{1,\ldots,K\}} \in \reals^{MK}\\
 A^\ast :=&\; (a_{m,m',k})_{m,m'\in\{1,\ldots,M\}, k \in \{1,\ldots,K\}} \in \reals^{M \times MK},
\end{align*} 
we have
$$
\log \lambda^\cont(\mathcal{X}_\tau) = \nu + A^\ast g^\cont(\mathcal{X}_\tau)
$$
which exhibits the same general form as \eqref{eq:model}. 

In order to formalize the connection between the multivariate Hawkes process in
\eqref{eq:hawkes} and the SEPP in \eqref{eq:model} we first describe our sampling process and the Hawkes and Poisson log likelihoods needed to prove Proposition \ref{prop:hawkes_formal}: The Hawkes process can be discretized by sampling $\lambda^{(c)}(\mathcal{X}_\tau)$ at $\tau=t\triangle$ for some sampling period $\triangle >0$ and letting 
\begin{equation} \label{eq:eleven} X_{t,m}=\sum_{i=N_{(t-1)\triangle}+1}^{N_{t\triangle}} \mathbb{I}_{m=m_i}\end{equation} 
for $t=1,\ldots, T$.  

Here $\mathbb{I}_E$ is the indicator function which returns 1 if $E$ is true and 0 if $E$ is false and $X_{t,m}$ is the number of events on node $m$ during the sampling interval $[(t-1)\triangle, t\triangle)$.   Overloading notation somewhat, let $\mathcal{X}_t=\{X_{s,m}\}_{s=1,\ldots, t m=1,\ldots,M}$ be the history of event counts up to time $t$.  The log-likelihood of the original Hawkes process observations up to time $T\triangle$ is \begin{align*}\ell_H(\mathcal{X}_{T\triangle}|\{\lambda_m^{(c)}\}_m)&=\sum_{i=1}^{N_{T\triangle}} \log \lambda_{m_i}^{(c)}(\tau_i)-\sum_{m=1}^M \int_0^{T\triangle} \lambda_m^{(c)} (\tau) d\tau\\ &=\sum_{m=1}^M \sum_{i=1}^{N_{T\triangle_{m.}}}  \log \lambda_{m_i}^{(c)}(\tau_i)- \int_0^{T\triangle} \lambda_m^{(c)} (\tau) d\tau.\end{align*}
Further note that if $X_{t,m} \sim \mbox{Poisson}(\lambda_{t,m})$, then the Poisson log likelihood is proportional to 
$$\ell_p(\mathcal{X}_T)|\{\lambda_m(\mathcal{X}_t)\}_{t,m})=\sum_{m=1}^M \sum_{t=1}^T [X_{t,m}\log(\lambda_m(\mathcal{X}_t))-\lambda_m(\mathcal{X}_t)].$$
We consider in Proposition \ref{prop:hawkes_formal} a SEPP with the intensity \begin{equation} \label{eq:fourteen} \lambda_{t,m}=\triangle \lambda_m^{(c)}(\mathcal{X}_{\triangle t}) \equiv \triangle \lambda_m^{(c)} (\triangle t; \mathcal{X}_{\triangle t}),\end{equation} where the last equality makes the sampling time explicit.  We now present a proposition which formalizes the connections between the SEPP and the log-linear Hawkes process.

\begin{prop} \label{prop:hawkes_formal} 
The likelihood of the discretized multivariate Hawkes data in \eqref{eq:eleven} can be approximated by the likelihood of the Poisson autoregressive model \eqref{eq:model} with intensity \eqref{eq:fourteen}, modulo terms independent of the unknown $\lambda^{(c)}$, where the approximation error depends on the sampling period $\triangle$. 
\end{prop}

This proposition suggests that the models and analysis we develop for SEPPs also provides insight into related Hawkes process models provided that the sampling period $\Delta$ is sufficiently small.

\section{Conclusion}

The proposed saturated SEPP allows
us to analyze statistical learning rates for a large class of point
processes, including discretized Hawkes processes, with long-range memory and saturation or clipping effects
common in real-world systems. The analysis presented in this paper
addresses instability issues present in prior works and incorporates 
a wide variety of structural assumptions on the ground truth processes
by allowing for arbitrary decomposable regularizers. The proposed bounds provide novel insight not only into sample complexity bounds, but also into phase transition boundaries dictated by stability and saturation effects that are supported by simulation results. In addition, experiments on data from neuroscience, criminology, and social media suggest that the models considered in this paper exhibit sufficient complexity to model real-world phenomena.

\bibliographystyle{plain}

\bibliography{final_version}

\appendix

\section{Proofs}

\subsection{Proof of Theorem \ref{Thm:bound}}

Theorem \ref{Thm:bound} is the combination of results from \cite{Neg10} and \cite{hall2016inference}.  We give a proof for the sake of completeness but claim no originality of techniques.  
For the first part of the proof, we follow Theorem 1 from \cite{Neg10}.  
By the definition of $\widehat{A}$ and properties of Bregman divergence for strongly convex functions, we have 
$$ \frac{R_{\min}}{2T} \sum_m \sum_t (\triangle_{m.}^\top g(\mathcal{X}_t))^2 \leq \frac{1}{T} \left|\sum_m \sum_t \epsilon_{t,m} \triangle_{m.}^\top g(\mathcal{X}_t)\right|+\lambda (\|A^\ast\|_{\mathcal{R}} -\|\widehat{A}\|_\mathcal{R}). \label{eq:sixteen}$$
where $R_{\min}$ is a strong convexity parameter for $e^x$ on the domain $x \in [R_{\min},R_{\max}]$.  Next note that
\begin{align*} \sum_m \sum_t \epsilon_{t,m} \triangle_{m.}^\top g(\mathcal{X}_t)&=\sum_m \sum_{m'} \triangle_{m,m'} \sum_t X_{t,m'} \epsilon_{t,m}\\
&=\langle \triangle , \sum_t \epsilon_t g(\mathcal{X}_t)^\top \rangle \\
&\leq \|\triangle\|_\mathcal{R} \|\sum_t \epsilon_t g(\mathcal{X}_t)^\top\|_{\mathcal{R}^\ast}. 
\end{align*}
Thus, assuming $\lambda/2>\frac{1}{T}\|\sum_t \epsilon_t g(\mathcal{X}_t)^\top\|_{\mathcal{R}^\ast}$ we have 
 \begin{equation} \label{eq: one} \frac{1}{T} \left|\sum_m \sum_t \epsilon_{t,m} \triangle_{m.}^\top g(\mathcal{X}_t)\right|+\lambda (\|A^\ast\|_{\mathcal{R}} -\|\widehat{A}\|_\mathcal{R}) \leq \frac{\lambda}{2} \|\triangle\|_\mathcal{R} +\lambda \|A^\ast\|_\mathcal{R}-\lambda \|\widehat{A}\|_\mathcal{R}. \end{equation}
Then
$$\|\widehat{A}\|_\mathcal{R}=\|A^\ast+\triangle\|_\mathcal{R}=\|A^\ast+\triangle_{\overline{\mathcal{M}}^\perp}+\triangle_{\overline{\mathcal{M}}}\|_\mathcal{R}.$$
Since $\mathcal{R}$ is decomposable with respect to the subspaces $(\mathcal{M},\overline{\mathcal{M}}^\perp)$, we have 
$$\|\widehat{A}\|_\mathcal{R} \geq \|A^\ast\|_\mathcal{R} +\|\triangle_{\overline{M}^\perp}\|_\mathcal{R}-\|\triangle_{\overline{\cal M}}\|_\mathcal{R}.$$ 
Thus 
\begin{equation}\frac{\lambda}{2} \|\triangle\|_\mathcal{R} +\lambda \|A^\ast\|_\mathcal{R}-\lambda \|\widehat{A}\|_\mathcal{R}\leq \frac{3\lambda}{2} \|\triangle_{\overline{\mathcal{M}}}\|_\mathcal{R} -\frac{\lambda}{2} \|\triangle_{\overline{\mathcal{M}}^\perp}\|_\mathcal{R} \leq \frac{3\lambda}{2} \|\triangle_{\overline{\mathcal{M}}}\|_\mathcal{R}.\label{eq:seventeen} \end{equation}
Recalling that $\Psi(\overline{\cal M})$ is the subspace compatibility constant, we have
\begin{equation*}\|\triangle_{\overline{\mathcal{M}}}\|_\mathcal{R} \leq \Psi(\overline{\mathcal{M}}) \|\triangle_{\overline{\mathcal{M}}}\|_F \leq \Psi(\overline{\mathcal{M}})\|\triangle\|_F.\end{equation*} 
It follows that 
$$ \frac{\lambda}{2} \|\triangle\|_\mathcal{R} +\lambda \|A^\ast\|_\mathcal{R}-\lambda \|\widehat{A}\|_\mathcal{R}  \leq \frac{3\lambda}{2} \Psi(\overline{\mathcal{M}}) \|\triangle\|_F.$$ 
Let $\|\triangle\|_T^2=\frac{1}{T} \sum_m \sum_t (\triangle_{m.}^\top g(\mathcal{X}_t))^2$ and therefore 
$$\|\triangle\|_T^2 \leq   \frac{3\lambda}{R_{\min}} \Psi(\overline{\mathcal{M}}) \|\triangle\|_F.$$
From here, we reduce the lower bound into the restricted eigenvalue condition.  Denote the subsets 
$$\mathcal{B}_\mathcal{R}=\{B \in \mathbb{R}^{M \times MK} : \|B_{\overline{\mathcal{M}}^\perp}\|_\mathcal{R} \leq 3 \|B_{\overline{\mathcal{M}}}\|_\mathcal{R}\}$$
and 
$$\mathcal{B}'_{\mathcal{R}}=\{B \in \mathcal{B}_\mathcal{R} : \|B\|_F=1\}.$$ 
Note that Equation \eqref{eq:seventeen} implies that $\triangle \in \mathcal{B}_{\cal R}$. Let $\mathcal{T}=\{p,2p,\ldots ,T\}$ so $|\mathcal{T}|/|T|=\frac{1}{p}$ (here we assume $\frac{T}{p}$ is an integer for simplicity). By Assumption \ref{A2} we have 
\begin{align*}
    \|\triangle\|_T^2 \geq& \frac{1}{T} \sum_{t\in \mathcal{T}} \sum_m \triangle_{m.}^\top \mathbb{E}[g(\mathcal{X}_t)g(\mathcal{X}_t)^\top |\mathcal{X}_{t-p}] \triangle_{m.} \\
    &\quad - \sup_{B \in \mathcal{B}_\mathcal{R}} \frac{1}{T} \sum_{t\in \mathcal{T}} \sum_m (b_{m.}^\top g(\mathcal{X}_t))^2 - \mathbb{E}[(b_{m.}^\top g(\mathcal{X}_t))^2 |\mathcal{X}_{t-p}]\\
 \geq& \frac{\omega}{p}\|\triangle\|_F^2  -  \sup_{B \in \mathcal{B}_\mathcal{R}} \frac{1}{T} \sum_{t\in \mathcal{T}} \sum_m (b_{m.}^\top g(\mathcal{X}_t))^2 - \mathbb{E}[(b_{m.}^\top g(\mathcal{X}_t))^2 |\mathcal{X}_{t-p}].
\end{align*}
We want to show 
\begin{equation}\label{eq:tech2} \sup_{B \in \mathcal{B}_\mathcal{R}} \frac{1}{T} \sum_{t\in \mathcal{T}} \sum_m (b_{m.}^\top g(\mathcal{X}_t))^2 - \mathbb{E}[(b_{m.}^\top g(\mathcal{X}_t))^2 |\mathcal{X}_{t-p}] \leq \frac{\omega\|\triangle\|_F^2}{2p}\end{equation}
with high probability, and we note that it suffices to show this for all $B \in \mathcal{B}'_{\mathcal{R}}$.

Now we define the matrix $G \in \mathbb{R}^{M \times M}$ as follows:
$$
G := \frac{1}{T}\sum_{t \in \mathcal{T}}(g(\mathcal{X}_t)g(\mathcal{X}_t)^T - \mathbb{E}[g(\mathcal{X}_t)g(\mathcal{X}_t)^T|\mathcal{X}_{t-p}]).
$$
Following the definition of $G$, and denoting each entry $G_{m,m'}$,
\begin{align*}
 \sup_{B \in \mathcal{B}'_{\mathcal{R}}} \frac{1}{T} \sum_t \sum_m (b_{m.}^\top  g(\mathcal{X}_t))^2 - \mathbb{E}\left[(b_{m.}^\top g(\mathcal{X}_t))^2 |\mathcal{X}_{t-p}\right] & = \sup_{B \in \mathcal{B}'_{\mathcal{R}}}\sum_{m=1}^M{b_{m.}^\top G b_{m.}}\\ & \leq \sup_{B \in \mathcal{B}'_{\mathcal{R}}}\|B\|_{2,1}^2 \max_{m,m’}|G_{m,m'}|.
\end{align*}
Recall that $\sup_{B \in \mathcal{B}'_{\mathcal{R}}}\|B\|_{2,1}^2 \leq \mu_{\mathcal{R}}$ by Assumption~\ref{A4}. Hence
\begin{align*}
 \sup_{B \in \mathcal{B}'_{\mathcal{R}}} \frac{1}{T} \sum_t \sum_m & (b_{m.}^\top  g(\mathcal{X}_t))^2 - \mathbb{E}\left[(b_{m.}^\top g(\mathcal{X}_t))^2 |\mathcal{X}_{t-p}\right] \leq \mu_{\mathcal{R}} \max_{m,m'}|G_{m,m'}|.
\end{align*}
Note that each entry $G_{m, m'}$ is a martingale and $|G_{m, m'}|\leq 2U^2$. Therefore we can apply the Azuma-Hoeffding inequality~\cite{Azuma}. For completeness, we state the Azuma-Hoeffding inequality as Theorem~\ref{Thm:supp4} in Section~\ref{sec:supplemental}. If we let
$$
Y_n := \frac{1}{T} \sum_{t =0}^n(g(\mathcal{X}_t)g(\mathcal{X}_t)^T - \mathbb{E}[g(\mathcal{X}_t)g(\mathcal{X}_t)^T|\mathcal{X}_{t-p}]),
$$
where $n = 0,1,2,..,|\mathcal{T}|$, and we set $t=\frac{\omega}{2\mu_{\mathcal{R}}p}$ and $c_n = \frac{2U^2}{T}$ as in Theorem~\ref{Thm:supp4} in Appendix~\ref{sec:supplemental}, we have 
$$
\mathbb{P}\left(|G_{m,m'}| \geq \frac{\omega}{2\mu_{\mathcal{R}}p}\right) \leq 2\exp\biggr( -\frac{T\omega^2}{32U^4p^2\mu_{\mathcal{R}}^2}\biggr).
$$
Applying a union bound,
$$
\mathbb{P}\left(\max_{m,m'}|G_{m,m'}| \geq \frac{\omega}{2\mu_{\mathcal{R}}p}\right) \leq 2M^2\exp\biggr(-\frac{T\omega^2}{32U^4p^2\mu_{\mathcal{R}}^2}\biggr).
$$
Hence if we set 
$$
T > \frac{128 U^4p^2 \mu_{\mathcal{R}}^2 \log M}{\omega^2},
$$
\eqref{eq:tech2} holds with probability at least $$1-\frac{2}{M^2},$$ guaranteeing that $\|\triangle\|_T^2 \geq \frac{\omega}{2p}\|\triangle\|_F^2$ with this same probability.  Putting everything together, we have
$$\|\triangle\|_F^2 \leq \frac{36p\Psi(\overline{\mathcal{M}})^2 \lambda^2}{R_{\min}^2 \omega^2}$$
with probability at least $1-\frac{2}{M^2}$ for $$T \geq \frac{128p^2U^4\mu_{\mathcal{R}}^2 \log M}{\omega^2}.$$  \qed

\subsection{Proof of Lemma \ref{Lem:ar1}}
Before we prove Lemma~\ref{Lem:ar1}, we first need the following supporting lemma.

\begin{lemma} \label{lem:cov_bound}
Let $Z=\min(\left\lfloor{\lambda}\right\rfloor,\tilde{U})$.  Define the random variables $\overline{X} \sim \mbox{Poisson}(\lambda)$, $X=\min(\overline{X},\tilde{U})$ and 
\[ Y=\begin{cases} 
      0 & \text{if   } X \leq Z \\
      1 & \text{if   } X > Z \\
      
   \end{cases}.
\] Then $$\mbox{Var}(Y) \leq \mbox{Var}(X).$$

\end{lemma}
\paragraph{Proof}We write $$X=(Y+Z)+(X-Y-Z).$$ Since $Z$ is a constant we have $$\mbox{Var}(X)=\mbox{Var}(Y)+\mbox{Var}(X-Y)+2\mbox{Cov}(Y,X-Y)$$ and it suffices to show $\mbox{Cov}(Y,X-Y) \geq 0$.  Conditioning on $Y$ gives \begin{align}\begin{split} \mathbb{E}\big[(Y-\mathbb{E}[Y])&(X-Y-\mathbb{E}[X-Y])\big]=\\ &p(Y=1)(1-\mathbb{E}[Y]) \mathbb{E}_{X,Y|Y=1}\big[X-Y-\mathbb{E}[X-Y]\big]\\
 +&p(Y=0)(-\mathbb{E}[Y]) \mathbb{E}_{X,Y|Y=0}\big[X-Y-\mathbb{E}[X-Y]\big].\label{eq:cov_expression}\end{split}\end{align}
Now observe that $Y=1$ implies $X-Y \geq Z$ (where we rely on the fact that $X$ and $Z$ are both integers) while $Y=0$ implies $X-Y \leq Z$; then
\begin{align}
    \mathbb{E}[X-Y|Y=1] \geq \mathbb{E}[X-Y]\label{eq:helper}\end{align} and \begin{align} \mathbb{E}[X-Y|Y=0] \leq \mathbb{E}[X-Y].
    \label{eq:helper2}
\end{align}
We argue that both terms in the sum in \eqref{eq:cov_expression} are non-negative.  For the first term, we have 
\begin{align*}
     \mathbb{E}_{X,Y|Y=1}\big[X-Y-\mathbb{E}[X-Y]\big]
    =\mathbb{E}_{X,Y|Y=1}[X-Y]-\mathbb{E}[X-Y] \geq 0
\end{align*}
by \eqref{eq:helper}.  
\begin{align*}
     \mathbb{E}_{X,Y|Y=0}\big[X-Y-\mathbb{E}[X-Y]\big]
    =\mathbb{E}_{X,Y|Y=0}[X-Y]-\mathbb{E}[X-Y] \leq 0
\end{align*} by \eqref{eq:helper2}.  Finally note that $\mathbb{E}[Y] \in (0,1)$ so that both terms in \eqref{eq:cov_expression} are indeed non-negative, and therefore $\mbox{Cov}(Y,X-Y) \geq 0$ as claimed. \qed 
~\newline

We now prove Lemma~\ref{Lem:ar1}.
Note that $$\mbox{Cov}(\min(X_{t,m},\tilde U),\min(X_{t,m'},\tilde U)|\mathcal{X}_{t-1}))=0$$ for $m \not =m'$.  We have 
\begin{align*}\mathbb{E}[\min&(X_t,\tilde U)\min(X_t,\tilde U)^\top|\mathcal{X}_{t-1}]= \\&\mathbb{E}[\min(X_t,\tilde U)|\mathcal{X}_{t-1}]\mathbb{E}[\min(X_t,\tilde U)|\mathcal{X}_{t-1}]^\top+\mbox{Diag}(\mbox{Var}(\min(X_t,\tilde U)|\mathcal{X}_{t-1}))
\end{align*}
where the first matrix is positive semi-definite because it is the outer product of a vector with itself.  Thus, to come up with a lower bound for our original matrix, we just need to lower bound the smallest element of $\mbox{Var}(\min(X_t,\tilde U)|\mathcal{X}_{t-1})$.   This amounts to lower bounding the variance of $\min(X_\lambda,\tilde U)$ where $X_\lambda$ is a Poisson random variable with mean $\lambda \in [R_{\min},R_{\max}]$.
Define
\[ Y_\lambda=\begin{cases} 
      0 & \text{if   } \min(X_\lambda,\tilde U) \leq \min(\left \lfloor{\lambda}\right \rfloor, \tilde U) \\
      1 & \text{if   } \min(X_\lambda,\tilde U) > \min(\left \lfloor{\lambda}\right \rfloor, \tilde U). \\
      
   \end{cases}
\]
By Lemma \ref{lem:cov_bound}, $\mbox{Var}(\min(X_\lambda, \tilde{U})) \geq \mbox{Var}(Y_\lambda)$ so our problem reduces to lower bounding the variance of $Y_\lambda$.  We argue that $$\mbox{Var}(Y_\lambda) \geq \min(\mbox{Var}(Y_{R_{\min}}),\mbox{Var}(Y_{R_{\max}}))$$ 
by considering two cases.  When analyzing these cases, we use the fact that $\mbox{Var}(Y_\lambda)$ will be minimized when the probability of outcome $(0)$ is either maximized or minimized.  We take $R_{\min} \leq \frac{1}{5}$ to make the exposition clearer.  At the end of the proof we discuss the $R_{\min}>\frac{1}{5}$ scenario which is virtually identical.

\paragraph{Case 1: $\lambda \in [R_{\min},\tilde U)$ where $\tilde U$ may be either greater than or less than $R_{\max}$}
\mbox{} \newline \indent In this scenario
\[ Y_\lambda=\begin{cases} 
      0 & \text{if   } X_\lambda \leq \left \lfloor{\lambda}\right \rfloor \\
      1 & \text{if   } X_\lambda > \left \lfloor{\lambda}\right \rfloor. \\
      
   \end{cases}
\]
 We claim $\mbox{Var}(Y_{R_{\min}}) \leq \mbox{Var}(Y_\lambda)$ for $\lambda \in [R_{\min},\tilde U)$.  To do this, we look at two subcases. 

First, if $1 \leq \lambda \leq \tilde U$, then basic properties of the median of the Poisson distribution imply that the probability of outcome $(0)$ will be between $\frac{1}{5}$ and $\frac{4}{5}$ and so $\mbox{Var}(Y_\lambda) \geq \frac{4}{25}$.  For the second case where $R_{\min} \leq \lambda<1$, outcome $(0)$ corresponds to
$$\mathbb{P}\left(X = 0 | X \sim \mbox{Poisson}(\lambda)\right)=\exp(-\lambda) \leq \exp(-R_{\min}).$$
Since $R_{\min} \leq \frac{1}{5}$ we get that $\exp(-R_{\min}) > \frac{4}{5}$.  Combining the two cases, we have concluded that $\mbox{Var}(Y_\lambda)$ is minimized on $\lambda \in [R_{\min},\tilde U]$ at $\lambda=R_{\min}$. Now, when $\lambda=R_{\min}$,  $$\mbox{Var}(Y_\lambda)=\exp(-R_{\min})(1-\exp(-R_{\min})).$$  Since $$f(x)=\frac{e^{-x}(1-e^{-x})}{x}$$ decreases monotonically on the interval $(0,\frac{1}{5}]$, we have $$\min_{x \in (0,\frac{1}{5}]} f(x) \geq f\left(\frac{1}{5}\right) \geq \frac{1}{2}.$$ 
Using the fact that $R_{\min} \in (0,\frac{1}{5}]$ we conclude   \begin{equation}\mbox{Var}(Y_{R_{\min}})=\exp(-R_{\min})(1-\exp(-R_{\min})) \geq \frac{1}{2}R_{\min}.\label{eq:bound1}\end{equation}

\paragraph{Case 2: $\lambda \in [\tilde U,R_{\max}]$}
\mbox{} \newline \indent Next we consider the variance when $\lambda \geq \tilde U$.  By the same reasoning as in Case 1, outcome $(0)$ can have probability no larger than $\frac{4}{5}$.  It remains to consider when outcome $(1)$ can have probability larger than $\frac{4}{5}$.  It is clear that for $\lambda \in [\tilde U,R_{\max}]$, outcome $(1)$ will be maximized for $\lambda=R_{\max}$.  When $\lambda=R_{\max}$, we directly compute the variance as
\begin{equation}\mbox{Var}(Y_{R_{\max}})=\sum_{i=0}^{\tilde{U}-1} \frac{R_{\max}^i \exp^{-R_{\max}}}{i!}\left(1-\sum_{i=0}^{\tilde{U}-1} \frac{R_{\max}^i \exp^{-R_{\max}}}{i!}\right)=\kappa. \label{eq:bound2}\end{equation}
Combining Case 1 and Case 2, we get that $$\mbox{Var}(Y_\lambda) \geq \min\left(\mbox{Var}(R_{\min}),\mbox{Var}(R_{\max})\right)$$ and combining Equations \eqref{eq:bound1} and \eqref{eq:bound2} gives the final result.

\paragraph{}If $R_{\min} \geq \frac{1}{5}$ an identical argument shows that for Case $1$ where $\lambda \in [R_{\min}, \tilde{U})$, $\mbox{Var}(Y_\lambda) \geq \frac{4}{25}$ and for Case 2 where $\lambda \in [\tilde U,R_{\max}]$, $\mbox{Var}(Y_\lambda) \geq \kappa$.  Hence, a lower bound on $\mbox{Var}(Y_\lambda)$ covering all possible values of $R_{\min}$ would be $\min(\frac{1}{2}R_{\min}, \kappa, \frac{4}{25})$.  In main body of the paper we present the bound for the $R_{\min} \leq \frac{1}{5}$ scenario in order to make the statement more interpretable. 

\qed

\subsection{Proof of Theorem \ref{Lem:two_basis}}
Recall that the AR$(2)$ model is a special case of  \eqref{eq:model} with $\phi_1[t]=\ind{t=1}$ and $\phi_2[t]=\ind{t=2}$.  With these choices of basis functions, 

$$g(\mathcal{X}_t)=[\min(X_1,\tilde U),\min(X_2,\tilde U)]^\top.$$
A computation shows that if we choose to condition on $\mathcal{X}_{t-1}$ as in the proof of Lemma \ref{Lem:ar1} we get a singular matrix.  However, Assumption \ref{A2} allows us to condition on $\mathcal{X}_{t-p}$ for any $p>0$ and so for this example it will be easiest to condition on $\mathcal{X}_{t-2}$.  We have
\begin{align*}\mathbb{E}[g(\mathcal{X}_t)g(\mathcal{X}_t)^\top|\mathcal{X}_{t-2}] =\mathbb{E}[g(\mathcal{X}_t)|\mathcal{X}_{t-2}]\mathbb{E}[g(\mathcal{X}_t)|\mathcal{X}_{t-2}]^\top+ \mbox{Cov}(g(\mathcal{X}_t)|\mathcal{X}_{t-2}).\end{align*}  
The first matrix is an outer product of a vector with itself so it is positive semi-definite and it suffices to lower bound the eigenvalues of the covariance matrix $\mbox{Cov}(g(\mathcal{X}_t)|\mathcal{X}_{t-2})$.  
Recall that a matrix $B$ is said to be strictly diagonally dominant if $$b_{i,i} -\sum_{j \not = i} |b_{i,j}|\geq \omega>0$$ for all $i$, and the eigenvalues of a symmetric strictly diagonally dominant matrix are lower bounded by $\omega$.  
To lower bound the eigenvalues of the covariance matrix, we will show it is strictly diagonally dominant.  We break the rows up into two cases.  The first case corresponds to rows where the diagonal depends on a lagged count $X_{t-1,m}$ while the second case corresponds to rows where diagonal depends on a count $X_{t,m}$ without a lag. 

\paragraph{Case 1: Rows $1$ through $M$}
The first $M$ rows of $\mbox{Cov}(g(\mathcal{X}_t)|\mathcal{X}_{t-2})$ have their diagonal of the form $\mbox{Var}(X_{t-1,m}|\mathcal{X}_{t-2}) \geq R_{\min}$.  We have $\mbox{Cov}(X_{t-1,m},X_{t-1,m'}|\mathcal{X}_{t-2})=0$ for all $m' \not =m$.  If node $m$ is not a parent of $m'$, then $X_{t-1,m}$ and $X_{t,m'}$ are independent conditioned on ${\cal X}_{t-2}$, so $\mbox{Cov}(X_{t-1,m},X_{t,m'}|\mathcal{X}_{t-2})=0$.  All that remains is to control $\mbox{Cov}(X_{t-1,m},X_{t,m'}|\mathcal{X}_{t-2})$ for the $\rho_{m}^{(c)}$ children of $m$.  

To do this, recall the decomposition $X_{t,m'}=\exp(\nu_{m'} +a_{m.'}^\top g(\mathcal{X}_{t-1}))+\epsilon_{t,m'}$.  For the remainder of the proof we let   $f_{m'}(\mathcal{X}_{t})=\exp(\nu_{m'} +a_{m.'}^\top g(\mathcal{X}_t))$ for notational simplicity and note that the Poisson noise term $\epsilon_{t,m}$ is zero mean conditioned on $X_{t-1,m}$.  Hence $$\mbox{Cov}(X_{t-1,m},X_{t,m'}|\mathcal{X}_{t-2})=\mbox{Cov}(X_{t-1,m},f_{m'}(\mathcal{X}_{t-1})|\mathcal{X}_{t-2}).$$  Since $f_{m'}(\mathcal{X}_{t-1})$ takes values in the interval $[R_{\min},R_{\max}]$, the variance of $f_{m'}(\mathcal{X}_{t-1})$ is bounded by a scaled Bernoulli random variable which takes values 0 with probability $\frac{1}{2}$ and $R_{\max}-R_{\min}$ with probability $\frac{1}{2}$.  This variance is equal to $\frac{(R_{\max}-R_{\min})^2}{4}$ and therefore 
\begin{align*}\mbox{Cov}(X_{t-1,m},f_{m'}(\mathcal{X}_{t-1})&|\mathcal{X}_{t-2}) \leq \\ & \sqrt{\mbox{Var}(X_{t-1,m}|\mathcal{X}_{t-2})\mbox{Var}(f_{m'}(\mathcal{X}_{t-1})|\mathcal{X}_{t-2})} \leq \frac{\sqrt{R_{\max}}(R_{\max}-R_{\min})}{2}.\end{align*}  
Hence the off diagonal entries sum to at most $$\frac{\rho_{m}^{(c)}\sqrt{R_{\max}}(R_{\max}-R_{\min})}{2}$$ so for these rows of the covariance matrix we have 
\begin{equation}\mbox{Cov}(g(\mathcal{X}_t))_{i,i}-\sum_{j \not =i}\mbox{Cov}(g(\mathcal{X}_t))_{i,j} \geq R_{\min}-\frac{\rho_{m}^{(c)}\sqrt{R_{\max}}(R_{\max}-R_{\min})}{2}.  \label{eq:first_bound}\end{equation}

\paragraph{Case 2: Rows $M+1$ through $2M$}
We next consider the final $M$ rows of the covariance matrix whose diagonal is of the form $\mbox{Var}(X_{t,m}|\mathcal{X}_{t-2}) \geq R_{\min}$.  We know $\mbox{Cov}(X_{t,m},X_{t-1,m'}|\mathcal{X}_{t-2})$ will be zero whenever node $m'$ is not a parent of $m$, and for the $\rho_{m}^{(p)}$ parents of $m$, the covariance is bounded below by $$\frac{\sqrt{R_{\max}}(R_{\max}-R_{\min})}{2}$$ just as in the previous paragraph.  

Finally, we need to consider $\mbox{Cov}(X_{t,m},X_{t,m'}|\mathcal{X}_{t-2})$.  When $m$ and $m'$ are not siblings this covariance will be zero.  When they do share a parent, we again recall the decomposition $X_{t,m}=f_m(\mathcal{X}_{t-1})+\epsilon_{t,m}$ and $X_{t,m'}=f_{m'}(\mathcal{X}_{t-1})+\epsilon_{t,m'}$ and note that the $\epsilon_{t,m}$ and $\epsilon_{t,m'}$ are zero mean conditioned on $X_{t,m'}$ and $X_{t,m}$ respectively.  Therefore 
$$\mbox{Cov}(X_{t,m},X_{t,m'}|\mathcal{X}_{t-2})=\mbox{Cov}(f_m(\mathcal{X}_{t-1}),f_{m'}(\mathcal{X}_{t-1})|\mathcal{X}_{t-2})$$ and using the fact that each $f_i(\mathcal{X}_t)$ takes values in the interval $[R_{\min},R_{\max}]$ it follows that this covariance is bounded by
$$\sqrt{\mbox{Var}(f_m(\mathcal{X}_{t-1})|\mathcal{X}_{t-2})\mbox{Var}(f_{m'}(\mathcal{X}_{t-1})|\mathcal{X}_{t-2})}\leq \frac{(R_{\max}-R_{\min})^2}{4}.$$  
 Recall that $\rho_{m}^{(s)}$ denotes the number of siblings of $m$.  Overall we have concluded that the sum of the off diagonal entries for the first $m$ rows is at most $$\frac{\rho_{m}^{(p)}\sqrt{R_{\max}}(R_{\max}-R_{\min})}{2}+\frac{\rho_{m}^{(s)}(R_{\max}-R_{\min})^2}{4}$$ so that for these rows we have \begin{align}\mbox{Cov}(g(\mathcal{X}_t))_{i,i}&-\sum_{j \not =i}\mbox{Cov}(g(\mathcal{X}_t))_{i,j}> \nonumber \\
 &R_{\min}-\frac{\rho_{m}^{(p)}\sqrt{R_{\max}}(R_{\max}-R_{\min})}{2}-\frac{\rho_{m}^{(s)}(R_{\max}-R_{\min})^2}{4}.\label{eq:second_bound}\end{align}
 We conclude that the smallest eigenvalue of the covariance matrix is lower bounded by the minimum of the two lower bounds in Equations \eqref{eq:first_bound} and \eqref{eq:second_bound}, and we define this minimum to be $r_\rho$. \qed

\subsection{Proof of Lemma \ref{Thm:Sparsity}}
We know from Equation \eqref{eq:seventeen} that $\|\triangle_{\mathcal{S}^\perp}\|_1 \leq 3 \|\triangle_\mathcal{S}\|_1$ and since $$\|\triangle\|_1=\|\triangle_{\mathcal{S}^\perp}\|_1+\|\triangle_\mathcal{S}\||_1$$ it follows that $\|\triangle\|_1 \leq 4\|\triangle_\mathcal{S}\|_1$. 
Recall that $\|v\|_1 \leq \sqrt{s} \|v\|_2$ for any s-sparse vector $v$.  Thus we have 
$$\|\triangle\|_1 \leq 4\|\triangle_\mathcal{S}\|_1 \leq 4\sqrt{s}\|\triangle_\mathcal{S}\|_F \leq 4\sqrt{s}\|\triangle\|_F.$$
For Assumption \ref{A5} we use a concentration result due to \cite{HoudreReynaudBouret03} in a similar manner as in \cite{hall2016inference}.  The result is restated as Theorem \ref{Thm:supp3} below.
Define $Y_n=\frac{1}{T} \sum_{t=1}^{n} g(\mathcal{X}_{t})_i\epsilon_{t,j}$ and note the following values
$$Y_n-Y_{n-1}=\frac{g(\mathcal{X}_{n})_i}{T}\epsilon_{n,j}$$ and $$M_n^k=\sum_{t=1}^n \mathbb{E}\left [\left(\frac{g(\mathcal{X}_{t})_i}{T}\epsilon_{t,j}\right)^k\big|\mathcal{X}_{t-1}\right ].$$
Thus $Y_n$ is a martingale.  We have $g(\mathcal{X}_{t})_i \leq U$, and by Lemma 1 in \cite{hall2016inference}, 
$$\epsilon_{t,j} \leq X_{t,j} \leq C \log(MT)$$
for all $t,j$ with probability at least $1-\exp(-cMT)$. Thus, 
$$|Y_n-Y_{n-1}| \leq \frac{CU\log(MT)}{T} =: B$$
with this same probability.
Next, note that 
\begin{align*}M_n^2&=\sum_{t=1}^n \mathbb{E}\left [\frac{g(\mathcal{X}_{t})_i^2}{T^2} \epsilon_{t,j}^2\big |\mathcal{X}_{t-1}\right] \\&\leq  \frac{1}{T^2} \sum_{t=1}^n U^2R_{\max}\\ &= \frac{n}{T^2} U^2R_{\max} =: \widehat{M_n}^2.\end{align*} 
Here we use that $\mathbb{E}[\epsilon_{t,j}^2|\mathcal{X}_{t-1}]$ is the variance of a Poisson random variable with mean bounded by $R_{\max}$, so it must also be bounded by $R_{\max}$.  
Next, we bound $M_n^k$:
$$M_n^k := \sum_{i=1}^n \mathbb{E}\left[\left(\frac{g(\mathcal{X}_i)_m}{T^2}(X_{i,l}-\mathbb{E}[X_{i,l}|\mathcal{X}_{i-1}])\right)^k\big|\mathcal{X}_{i-1}\right] \leq B^{k-2}M_n^2.$$
In the language of Theorem \ref{Thm:supp3}, 
\begin{align*}D_n &:= \sum_k \frac{\gamma^k}{k!} M_n^k \\& \leq  \frac{\widehat{M_n}^2}{B^2} \sum_k \frac{ \gamma^k B^k}{k!} =: \widehat{D}_n.\end{align*}
Let $\tilde{D_n}$ corresponds to the negative sequence of $D_n$, and so it is still bounded by $\widehat{D_n}$.  Using Markov's inequality, we get
\begin{align*}\mathbb{P}(|Y_n| \geq y)&=\mathbb{P}(Y_n \geq y) + \mathbb{P}(-Y_n \leq y)\\
&\leq \mathbb{E}[\exp(\gamma Y_n)] \exp(-\gamma y)+\mathbb{E}[\exp(-\gamma Y_n)]\exp(-\gamma y)\\& \leq \mathbb{E}[\exp(\gamma Y_n-D_n)]\exp(\widehat{D_n}-\gamma y) + \mathbb{E}[\exp(-\gamma Y_n)-\tilde{D_n}]\exp(\widehat{D_n}-\gamma y).\end{align*}
Using Theorem \ref{Thm:supp3} we conclude that 
$$\mathbb{E}[\exp(\gamma Y_n-D_n)] \leq 1 $$
and $\mathbb{E}[\exp(\gamma Y_n-\tilde{D_n})]\leq 1$ so
$$ \mathbb{P}(|Y_n| \geq y) \leq 2\exp(\widehat{D}_n-\gamma y).$$
We set
$$\gamma=\frac{1}{B} \log \left(1+\frac{yB}{\widehat{M_n}^2}\right)$$ 
and to simplify things note that $(1+x)\log(1+x)-x \geq \frac{3x^2}{2(x+3)}$.  Putting everything together gives
\begin{align*}\mathbb{P}(|Y_T \geq y|) &\leq 2\exp \left(\frac{-3y^2}{2yB+6\widehat{M_n}^2}\right)\\&=2\exp \left(\frac{3y^2T}{2UC\log(MT)y+6R_{\max}}\right).\end{align*}  Now, recall that 
$$\frac{\lambda}{2}=\frac{4CR_{\max}U^2\log^2(MT)}{\sqrt{T}}$$
and setting $y=\frac{\lambda}{2}$ gives 
\begin{align*}\mathbb{P}(Y_T \geq \frac{\lambda}{2}) &\leq 2\exp \left(\frac{48C^2U^4\log^4(MT)}{2C^2U^3\log^3(MT)/\sqrt{T}+6U^2R_{\max}}\right)\\&=2\exp \left(\frac{48U\log(MT)}{2/\sqrt{T}+\frac{6R_{\max}}{C^2U\log^3(MT)}}\right) \\&\leq 2\exp \left(\frac{48U\log(MT)}{8}\right).\end{align*}
Taking a union bound over all $i,j$ gives us 
\begin{align*}\mathbb{P}\Big(\max_{i,j} \frac{1}{T} \left|\sum_{t=1}^T g(\mathcal{X}_{t})_i \epsilon_{t,j}\right| \geq 4U^2&\log^2(MT)/\sqrt{T}\Big)\\ &\leq \exp(\log(2M^2)-6U\log(MT))\\ &\leq \exp(3\log(MT)-6U\log(MT))\\ &=\exp(-c\log(MT))\end{align*}
for $c=6U-3$ which is positive since $U \geq 1$.  
In the final statement of the proof we assume $C\log(MT) \geq U$ and replace $U$ with $C\log(MT)$ in order to limit the number of constants and make the crucial dependencies clear.  This assumption should hold for reasonable choices of $U$ in the settings we imagine in practice, but if not, a factor of $\log^2(MT)$ can be replaced by $U^2$ in the final bound.  \qed

\subsection{Proof of Lemma \ref{Thm:Group}}
For Assumption \ref{A3}, note that any $A \in \overline{\mathcal{S}_G}$ satisfies $A_{.c}=0$ for $i \not \in S_G$.  Therefore 
\begin{align*}\|\triangle\|_G \leq 4\|\triangle_{\overline{\cal S}_G}\|_G&=4\sum_{i \in S_G} \|\triangle_{.i}\|_2 \\ &\leq 4\sqrt{s_G} \|\triangle_{\overline{\cal S}_G}\|_F \\
&\leq 4 \sqrt{s_G}\|\triangle\|_F.\end{align*}  
For Assumption \ref{A4}, we have $\|\triangle\|_G^2 \leq 16s_G \|\triangle\|_F^2$ from the previous paragraph, and we claim $\|A\|_{2,1} \leq \|A\|_G$ for any matrix $A$.  To see this, we compute 
$$\|A\|_G^2=\left( \sum_c \sqrt{\sum_r a_{r,c}^2}\right)^2=\sum_c \sum_{c'} \sqrt{(\sum_r a_{r,c}^2)(\sum_r a_{r,c'}^2)}$$
 while 
 $$\|A\|_{2,1}^2=\sum_r (\sum_c |a_{r,c}|)^2=\sum_c \sum_{c'} \sum_r |a_{r,c}\|a_{r,c'}|.$$ To complete the proof, we fix $c,c'$ and need to show
 $$\sum_r |a_{r,c}\|a_{r,c'}| \leq \sqrt{(\sum_r a_{r,c}^2)(\sum_r a_{r,c'}^2)},$$
 or equivalently that 
 $$(\sum_r |a_{r,c}\|a_{r,c'}|)^2 \leq (\sum_r a_{r,c}^2)(\sum_r a_{r,c'}^2).$$  
 We have 
 $$(\sum_r |a_{r,c}\|a_{r,c'}|)^2=\sum_r \sum_{r'} |a_{r,c}a_{r,c'}a_{r',c}a_{r',c'}|.$$
 Let $\mathcal{J}$ denote all two element combinations of $M$ and we can write 
 $$\sum_r \sum_{r'} |a_{r,c}a_{r,c'}a_{r',c}a_{r',c'}|=\sum_r (a_{r,c}a_{r,c'})^2+\sum_{(i,j) \in \mathcal{J}} 2|a_{i,c}a_{j,c}a_{i,c'}a_{j,c'}|.$$ 
 On the other hand, 
\begin{align*}(\sum_r a_{r,c}^2)(\sum_r a_{r,c'}^2)&=\sum_r \sum_{r'} a_{r,c}^2 a_{r',c'}^2\\&=\sum_r (a_{r,c}a_{r,c'})^2+\sum_{(i,j) \in \mathcal{J}} (a_{i,c}a_{j,c'})^2 +(a_{j,c}a_{i,c'})^2.\end{align*}
The proof follows from noting that
$$(a_{i,c}a_{j,c'})^2 +(a_{j,c}a_{i,c'})^2 \geq 2|a_{i,c}a_{j,c}a_{i,c'}a_{j,c'}|$$ 
for any real numbers $a_{i,c},a_{i,c'},a_{j,c},a_{j,c'}$.

For Assumption \ref{A5}, we rely on Theorem 1 in \cite{rakh} which is restated as Theorem \ref{Thm:supp2}.  In our setup, we need to bound the $l_2$ norm of the $m$th column of $\frac{1}{T} \sum_t \epsilon_tg(\mathcal{X}_t)^\top$. 
Note that the $l_2$ norm is 2-smooth, because for any $x,y \in \mathbb{R}^m$ we have
\begin{align*}\|x+y\|^2+\|x-y\|^2&=\langle x+y,x+y\rangle+\langle x-y,x-y\rangle \\
&= 2\langle x,x\rangle+2\langle y,y\rangle.\end{align*}
In the language of Theorem \ref{Thm:supp2}, for a fixed $m$ we form a martingale difference sequence $\{Z_t\}$ with  $$Z_t=\frac{1}{T}\left( \epsilon_t g(\mathcal{X}_t)^\top \right)_{.m}$$ so that
$$\|Z_t\|_2=\frac{1}{T}\sqrt{\sum_{m'} (g(\mathcal{X}_t)_{m}\epsilon_{t,m'})^2}=\frac{g(\mathcal{X}_t)_m}{T} \sqrt{\sum_{m'} \epsilon_{t,m'}^2}.$$ 
We know $g(\mathcal{X}_t)_m \leq U$ and by Lemma 1 from \cite{hall2016inference} $\epsilon_{t,m'} \leq C \log(MT)$ with probability at least $1-\exp(-cMT)$.  We conclude 
$$\|Z_t\|_2 \leq \frac{1}{T}U \sqrt{MC\log^2(MT)}$$ and thus $$\sum_{t=1}^T \|Z_t\|_2^2 \leq CU^2\log^2(MT) \frac{M}{T}.$$
To compute the constant $Q_{\max}$ appearing in Theorem \ref{Thm:supp2} we let $R(x)=x^\top x$ so that $\nabla R (x)=x$.  Then for any $x,y$ in the unit ball with respect to the $\|\cdot \|_2$ norm, we have 
$$D_R(x,y)=\|x\|_2^2-\|y\|_2^2 -\langle y,x-y\rangle \leq \|x\|_2^2+\|y\|_2\|x-y\|_2 \leq 3.$$ 
by Cauchy-Schwarz.
Thus we can take $Q_{\max}=\sqrt{3}$.  To simplify, we note $W_n \leq V_n$ and
$$(\mathbb{E}[\sqrt{V_n+W_n}])^2 \leq \mathbb{E}[V_n+W_n] \leq 2V_n.$$ 
Further, $2.5Q_{\max}(\sqrt{V_n}+1) \leq 5\sqrt{V_n}$.  With these simplifications, Theorem 1 from \cite{rakh} says that
$$\mathbb{P}\left(\frac{1}{T}\left\|\big(\sum_t \epsilon_t g(\mathcal{X}_t)^\top \big)_{.m}\right\|_2 >(5+2u)V_n\right) \leq \sqrt{2}\exp \left(-\frac{u^2}{16}\right).$$
Setting $u=\log(T)$ and plugging in our values for $V_n$ we conclude that
$$\mathbb{P}\left(\frac{1}{T}\left\|\big(\sum_t \epsilon_t g(\mathcal{X}_t)^\top \big)_{.m} \right\|_{2}>CU \log^2(MT) \sqrt{\frac{M}{T}}\right) \leq \sqrt{2}{\exp(-\log^2(MT)}).$$ 
Taking a union bound over all $m$, we get that $$\frac{1}{T}\left\|\big(\sum_t \epsilon_t g(\mathcal{X}_t)^\top \big)_{.m}\right\|_{2} \leq CU \log^2(MT) \sqrt{\frac{M}{T}}$$ for all $m$ with probability at least $1-\sqrt{2}\exp(-\log(MT)) = 1 - \frac{\sqrt{2}}{MT}$.  \qed

\subsection{Proof of Lemma \ref{Thm:Rank}}
For Assumption \ref{A4}, from the statement of Lemma \ref{Thm:Rank} we have $\|A^\ast\|_{2,1}^2 \leq D\sqrt{M}$ and we search for $\widehat{A}$ over the ball $\{A: \|A\|_{2,1}^2 \leq D\sqrt{M}\}$.  Thus
$$\sup_{B \in \mathcal{B}'_{\mathcal{R}}} \|B\|_{2,1}^2 \leq 2D\sqrt{M} =\mu_\mathcal{R}.$$
In contrast to the sparsity case, Assumption \ref{A3} is nontrivial to verify
in the low-rank case because $\cal W \not =\overline{\cal W}$.  However, this condition was shown in Lemma 3.4 of \cite{rechtlemma}.

\sloppypar For Assumption \ref{A5} we rely on the notion of a $k$-regular normed vector space defined in Section \ref{sec:supplemental} as well as Theorem 2.1 from \cite{regularthm} which is stated in Theorem \ref{Thm:supp}.  Further, Example 3.1 in \cite{regularthm} establishes that $(\mathbb{R}^{M \times MK},\|\cdot\|_\ast)$ is $k$-regular for $k=3\log(\min(M,N))$.  In the language of Theorem \ref{Thm:supp} we form a martingale difference sequence $\{\zeta_t\}$ with 
$\zeta_t=\frac{1}{T}\epsilon_t g(\mathcal{X}_t)^\top$ and then $$\|\zeta_t\|_{op}=\frac{\epsilon_t^\top g(\mathcal{X}_t)}{T}=\frac{1}{T}\sum_{m=1}^M \epsilon_{t,m} g(\mathcal{X}_{t})_{m}.$$
Consider the random variable 
\begin{equation} \label{eq:lm1} \sum_{m=1}^M \epsilon_{t,m} g(\mathcal{X}_{t})_{m}.\end{equation}
We have $g(\mathcal{X}_t)_{m} \leq U$ and $\epsilon_{t,m} \leq C\log(MT)$ for all $t,m$ with probability at least $e^{-cMT}$ by Lemma 1 from \cite{hall2016inference}.  Further, conditioned on $\mathcal{X}_t$, the $\epsilon_{t,m}g(\mathcal{X}_{t})_{m}$ are all independent, so \eqref{eq:lm1} is a sum of zero mean independent random variables bounded by $CU\log(MT)$.  Hence, we have 
 $$\sum_{m=1}^M \mathbb{E}[\epsilon_{t,m}g(\mathcal{X}_{t})_{m}|\mathcal{X}_{m-1}] \leq CM\log^2(MT)U^2,$$  
 and applying Bernstein's inequality gives 
 $$\mathbb{P}\left(|\epsilon_t^\top g(\mathcal{X}_t)| > \sqrt{M}\log^2(MT)\right) \leq 2\exp \left(-\frac{\log^4(MT)/2}{C\log^2(MT)U^2+\frac{CU\log(MT)}{3\sqrt{M}}}\right).$$
 Therefore
 \begin{align*}\mathbb{P}\Big(|\epsilon_t^\top g(\mathcal{X}_t)|  > \sqrt{M}&\log^2(MT) \text{ for at least one t }\Big) \\ & \leq 2\exp \left(\log(T)-\log^4(MT))/(2C\log^2(MT)U^2+1)\right)\\ & \leq \exp \left(-\frac{\log^4(MT)}{4C\log^2(MT)U^2+2}\right).\end{align*}
  We apply Theorem \ref{Thm:supp} with $k=3\log(M)$, 
$$\sum_{i=1}^T \sigma_i^2=\frac{M\log^4(MT)}{T}$$ 
and $\gamma=\log(T)$.  This gives 
\begin{align*}\mathbb{P}\Big(\frac{1}{T} \|\sum_t \epsilon_tg(\mathcal{X}_t)^\top \|_{op} > ( 3\sqrt{2}\log(M)+&\sqrt{2}\log(T)) \log^4(MT) \sqrt{\frac{M}{T}}\Big) \\ & \leq \exp\left(\frac{-\log^2(T)}{2}\right).\end{align*} \qed

\subsection{Proof of Proposition \ref{prop:hawkes_formal}}

In this proof, we take $\lambda_m^{(c)}(\tau)$ to mean $\lambda_m^{(c)}(\tau;\mathcal{X}_\tau)$.  Using the approximation $$\int_{(t-1)\triangle}^{\triangle t} \lambda_m^{(c)}(\tau) d\tau \approx \triangle \lambda_m^{(c)} (\triangle t)$$ we derive an approximate sampled Hawkes (SH) log-likelihood proportional to 
$$ \ell_H (\mathcal{X}_{T\triangle}|\{\lambda_m^{(c)}\}_m) \approx \sum_{m=1}^M \sum_{t=1}^T [X_{t,m}\log \lambda_m^{(c)}(\triangle t) - \triangle \lambda_m^{(c)}(\triangle t)] =: \ell_{SH}(\mathcal{X}_T | \{\lambda_m^{(c)}\}_m).$$
If $X_{t,m}$ were generated according to \eqref{eq:model} with intensity \eqref{eq:fourteen} for $T=1,\ldots T$, then, ignoring terms independent of $A^\ast$, 
$$\ell_P(\mathcal{X}_t|\{\triangle \lambda_m^{(c)}(\mathcal{X}_{\triangle t})\}_{t,m}) \coloneqq \sum_{m=1}^M \sum_{t=1}^T [X_{t,m} \log \triangle \lambda_m^{(c)}(\triangle t)-\triangle \lambda_m^{(c)}(\triangle t)].$$ 
Note that $$\ell_P(\mathcal{X}_T |\{\triangle \lambda_m^{(c)}(\mathcal{X}_{\triangle t})\}_{t,m}=\ell_{SH}(\mathcal{X}_T|\{\lambda_m^{(c)}\}_m)+C$$ where the constant $C$ depends on $\triangle$ but is independent of $\lambda_m^{(c)}$. \qed

\subsection{Extension to more general saturation functions} In the main body of the paper the only saturation function we consider is $f(x)=\min(x,\tilde U)$ for purposes of simplicity, but our theory extends to a larger class of saturation functions $f$.  However, for our analysis it is crucial to assume that the function $f$ is bounded so that we can define the maximum and minimum rates, $R_{\min}$ and $R_{\max}$, from which each observation is drawn.  The only place where we rely on the structure of $f$ beyond its boundedness is in proving the restricted eigenvalue condition in Assumption \ref{A2}.  In the case of the ARMA$(1,1)$ model, we show our results extend to monotonic differentiable functions in Proposition \ref{prop:saturation} below. 

\begin{prop} \label{prop:saturation} Suppose $(X_t)_{t=1}^T$ is generated according to the ARMA$(1,1)$ model in  \eqref{eq:arma} with a general saturation function $f$ applied entrywise to the vector $X_t$.  Suppose $f$ is bounded on $\mathbb{R}$, monotonically increasing, and differentiable with $f'(x) \geq c$ on $[0,R_{\max}]$.  Then $$\lambda_{\min}[\mathbb{E}[g(\mathcal{X}_t)g(\mathcal{X}_t)^\top|\mathcal{X}_{t-1}] \geq c^2\min(\frac{1}{2}R_{\min},\kappa).$$
\end{prop}
\paragraph{Proof} We have \begin{align*} \mathbb{E}[g(\mathcal{X}_t)g(\mathcal{X}_t)^\top|\mathcal{X}_{t-1}]= \mathbb{E}[g(\mathcal{X}_t)|\mathcal{X}_{t-1}]\mathbb{E}[g(\mathcal{X}_t)|\mathcal{X}_{t-1}]^\top+\mbox{Diag}(\mbox{Var}(g(\mathcal{X}_t)|\mathcal{X}_{t-1}))\end{align*}
where the first matrix is positive semi-definite because it is the outer product of a vector with itself.  Thus, to come up with a lower bound for our original matrix, we just need to lower bound the smallest element of $\mbox{Var}(g(\mathcal{X}_t)|\mathcal{X}_{t-1})$.   This amounts to lower bounding the variance of $f(X)$ where $X$ is a Poisson random variable with mean $\lambda \in [R_{\min},R_{\max}]$.

Let $p=\mathbb{P}(X \leq \left \lfloor{\lambda}\right \rfloor)$ so $1-p=\mathbb{P}(X \geq \left \lceil{\lambda}\right \rceil)$.  Consider the random variable $X'$ which takes the value $\left \lfloor{\lambda}\right \rfloor$  with probability $p$ and $\left \lceil{\lambda}\right \rceil$  with probability $1-p$.  Since $f$ is monotonic, the argument from Lemma \ref{lem:cov_bound} shows that $\mbox{Var}(f(X')) \leq \mbox{Var}(f(X))$ so we reduce our problem to lower bounding the variance of $f(X')$.  

Note that this variance is equal to the shifted random variable $X''$ defined by $f(X')=0$ with probability $p$ and $f(\left \lceil{\lambda}\right \rceil)-f(\left \lfloor{\lambda}\right \rfloor)$ with probability $1-p$ which is a scaled Bernoulli random variable with variance 
$$(f(\left \lceil{\lambda}\right \rceil)-f(\left \lfloor{\lambda}\right \rfloor))^2p(1-p).$$ 
Since $f'(x) \geq c$ on $[0,R_{\max}]$,
$$f(\left \lceil{\lambda}\right \rceil)-f(\left \lfloor{\lambda}\right \rfloor) \geq c$$ 
and so the lower bound on our variance becomes $c^2p(1-p)$.  Finally, by Lemma \ref{Lem:ar1} we have $p(1-p) \geq \min(\frac{1}{2}R_{\min},\kappa)$ which completes the proof. \qed
\subsection{Supplemental Theorems}

\label{sec:supplemental}
\paragraph{Definitions} Before introducing martingale concentration results, we give the following definitions.

\begin{definition}A Banach space $(E,\|\cdot \|)$ is \textit{s-smooth} if there exists $C>0$ satisfying 
$$\|x+y\|^s+\|x-y\|^s \leq 2\|x\|^s+2C^s\|y\|^s$$ 
for all $x,y \in E.$
\end{definition} 

Note that $(\mathbb{R}^M, \|\cdot \|_2)$ is 2-smooth with $C=1$ because
$$\|x+y\|^2+\|x-y\|^2=\langle x+y,x+y\rangle+\langle x-y,x-y\rangle=2\langle x,x\rangle+2\langle y,y\rangle.$$

\begin{definition}A Banach space $(E,\|\cdot \|)$ is $k-\textit{regular}$ if there exists $k+ \in [1,k]$ along with a norm $\|\cdot \|_+$ such that $(E, \|\cdot \|_+)$ is $k_+$-smooth and 
$$\|x\|^2 \leq \|x\|_+^2 \leq \frac{k}{k_+} \|x\|^2$$ 
for all $x \in E$.  

\end{definition}

By Example 3.3 from \cite{regularthm}, the space $(\mathbb{R}^{M \times N},\|\cdot \|_\ast)$ is $k$-regular for $k=3\log(\min(M,N))$.  

\begin{theorem}
\label{Thm:supp} (Theorem 2.1.iii in \cite{regularthm})

Let $(E,\|\cdot\|)$ be $k$-regular and let $\zeta_i$ be an E-valued martingale difference sequence with $\|\zeta_i\| \leq \sigma_i$.  Let $S_N=\sum_{i=1}^N \zeta_i$.  Then $$\mathbb{P}\left(\|S_N\| \geq (\sqrt{2k} +\sqrt{2}\gamma) \sqrt{\sum_{i=1}^N \sigma_i^2}\right) \leq \exp(-\frac{\gamma^2}{2}).$$

\end{theorem}

\begin{theorem}
\label{Thm:supp2} (Theorem 1 in \cite{rakh})
Let $(E, \|\cdot \|)$ be a 2-smooth Banach space. Let R be a function which is 1-strongly convex on the unit ball in the dual norm of $\|\cdot \|$.  $$D_R: B_\ast \times B_\ast \to \mathbb{R}$$ be the Bregman divergence with respect to $R$, and finally let $Q_{\max}^2=\sup_{x,y \in B_\ast} D_R(f,g)$. Let $Z_1,\ldots , Z_n$ be a martingale difference sequence with $V_n=\sum_{t=1}^n \|Z_t\|^2$ and $W_n=\sum_{t=1}^n \mathbb{E}_{t-1}\|Z_t\|^2$.  Then 
\begin{align*}\mathbb{P}\Bigg(\|\sum_{t=1}^n Z_t\|> 2.5Q_{\max}(\sqrt{V_n}+1)+& u\sqrt{V_n+W_n+(\mathbb{E}[\sqrt{V_n+W_n}])^2}\Bigg)\\ &\leq \sqrt{2}\exp(-\frac{u^2}{16}).\end{align*}

\end{theorem}

\begin{theorem}
\label{Thm:supp3} (Lemma 3.3 in \cite{HoudreReynaudBouret03})
Let $(Y_n)$ be a martingale and let $$M_n^k=\sum_{i=1}^n \mathbb{E}[(Y_i-Y_{i-1})^k|\mathcal{Y}_{i-1}].$$ Let $\gamma$ be such that for all $i \leq n$, we have $$\mathbb{E}[\exp(|\gamma(Y_i-Y_{i-1})|)] \leq \infty.$$ Then $$\epsilon_n=\exp(\gamma Y_n-\sum_{k \geq 2} \frac{\gamma^k}{k!} M_n^K)$$ is a super-martingale.  Moreover, if $Y_0=0$ then $\mathbb{E}[\epsilon_n] \leq 1$.

\end{theorem}

\begin{theorem}
\label{Thm:supp4} (Azuma-Hoeffding inequality)
Let $(Y_n)$ be a martingale and $|Y_n - Y_{n-1}| < c_n$. Then 
$$\mathbb{P}(|Y_N - Y_0| \geq t) \leq 2\exp\biggr( -\frac{t^2}{2\sum_{n=1}^N {c_n^2}}\biggr).$$  
\end{theorem}

\end{document}